\pdfoutput=1
\documentclass[11pt]{article}
\usepackage{authblk}
\usepackage[final]{acl2023}
\usepackage{times}
\usepackage{xcolor}
\usepackage[thinlines]{easytable}
\usepackage{graphicx}
\usepackage{latexsym}
\usepackage{array}
\usepackage{booktabs}
\usepackage{pgfplots}
\usepackage{algorithm}
\usepackage{algpseudocode}
\usepackage{graphicx}
\usepackage[T1]{fontenc}
\usepackage[utf8]{inputenc}
\usepackage{microtype}
\usepackage{xcolor}
\usepackage{soul}
\usepackage{fullpage}
\usepackage{enumitem}
\usepackage{pgfplots}
\usepackage{algorithm}
\usepackage{algpseudocode}
\usepackage{graphicx}
\usepackage{placeins}
\usepackage{tabularx}
\usepackage{makecell}
\usepackage{booktabs}       % professional-quality tables
\usepackage{array}          % tables with fixed lengths - enables using "m" to center content of non-multirow cells
\usepackage{colortbl}
\usepackage{times}
\usepackage{graphicx}
\usepackage{latexsym}
\usepackage{array}
\usepackage{booktabs}
\usepackage{pgfplots}
\usepackage{algorithm}
\usepackage{algpseudocode}
\usepackage{graphicx}
\usepackage[T1]{fontenc}
\usepackage[utf8]{inputenc}
\usepackage{microtype}
\usepackage{xcolor}
\usepackage{soul}

\title{To Asymmetry and Beyond: Structured Pruning of Sequence to Sequence Models for Improved Inference Efficiency  \thanks{~~~Corresponding author: dcampos3@illinois.edu}}
\author[1,2]{Daniel Campos}
\author[1]{ChengXiang Zhai}
\affil[1]{Department of Computer Science, the University of Illinois Urbana-Champaign}
\affil[2]{Neeva Inc.}
\begin{document}
\maketitle
\begin{abstract}
Sequence-to-sequence language models can be used to produce abstractive summaries which are coherent, relevant, and concise. Still, model sizes can make deployment in latency-sensitive or web-scale implementations difficult. This paper studies the relationship between model size, structured pruning, inference efficiency, and summarization accuracy on widely used summarization datasets. We show that model accuracy is tied to the encoder size while inference efficiency is connected to the decoder. Using asymmetric pruning can lead to nearly 3x improvement in inference latency with ~1 point loss in Rouge-2. Moreover, we find both the average degradation and the role of asymmetry to be consistent across model sizes and variations in datasets.  We release our code\footnote{https://github.com/spacemanidol/Efficient-Web-Scale-Absractive-Summarization}, training regimes, and associated model \footnote{https://huggingface.co/spacemanidol} for broad usage to encourage usage and experimentation. 
 \end{abstract}
\begin{figure}[!htb]
\begin{tikzpicture}
\scalebox{0.85}{
\begin{axis}[
    title={Accuracy vs. Inference Speed},
    xlabel={Inference Speedup},
    ylabel={\% Degradation in Rouge-2},
    xmin=1, xmax=4,
    ymin=-40 , ymax=2,
    xtick={1,2,3},
    ytick={-30,-20,-10, 0},
    legend pos=south east,
    ymajorgrids=true,
    grid style=dashed,
    legend style={nodes={scale=0.8, transform shape}}, 
    legend image post style={}
]
\addplot[
    color=blue,
    mark=triangle,
    ]
    coordinates {
    (1.18, 0.59) (1.43, -2.16) (1.8,-4.36) (2.44, -5.99) (3.83, -9.85)
    };
\addplot[
    color=red,
    mark=square,
    ]
    coordinates {
    (1.01, -2.62) (1.03, -3.93) (1.04,-8.42) (1.04, -10.57) (1.06, -18.89)
    };
\addplot[
    color=green,
    mark=star,
    ]
    coordinates {
    (1.19, -2.80) (1.40, -3.71) (1.91,-11.69) (2.20, -17.62) (2.44, -39.06)
    };
\legend{Prune Decoder, Prune Encoder, Prune Both}
 \end{axis}}
\end{tikzpicture}
    \centering
    \caption{Impact of Asymmetrical Pruning on inference speedups and ROUGE-2 degradation on Query Independent Web Summarization. Inference Time is the mean inference time for a batch size of 1 on an A10 GPU over seven iterations.   }
    \label{fig:speed}
\end{figure}
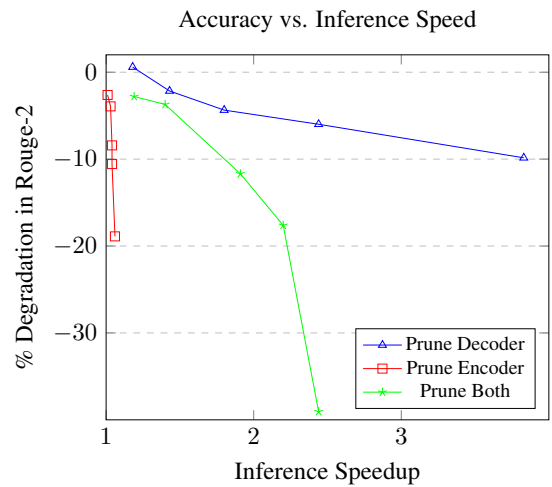
\section{Introduction}
The application of sequence-to-sequence language models has become an important tool for natural language processing tasks such as machine translation \cite{Sutskever2014SequenceTS}, audio transcription \cite{Radford2022RobustSR}, and abstractive summarization \cite{Raffel2020ExploringTL}. Sequence-to-sequence models effectively turn each of these aforementioned tasks into two-step problems: extraction and generation, and heavily condition the generation on the input. \\
Besides ensuring on-topic responses sequence to sequence models have the added benefit of being able to map inputs to targets with varying lengths and modalities in ways encoder or decoder-only systems cannot. \\
When used for abstractive summarization, sequence-to-sequence modeling has two steps, extraction using the encoder and generation using the decoder, which usually involves repeated execution until an end-of-sequence token is emitted. Since the encoder runs once on the input \cite{Sutskever2014SequenceTS} its cost of execution is proportional to the batch size. The cost of decoder execution can be highly variable based on the generation length \cite{Tay2021ScaleEI}. Despite the broad study of sequence-to-sequence models \cite{Raffel2020ExploringTL} and how they compress \cite{Li2022DQBARTES}, the role of model symmetry as applied to inference efficiency and model accuracy is lacking. \\
Recent advances in scaling language models have led to a wide study on \textit{scaling laws} as applied to language model performance \cite{Kaplan2020ScalingLF}, training data size \cite{Hoffmann2022TrainingCL}, machine translation \cite{Henighan2020ScalingLF}, and even reinforcement learning \cite{Neumann2022ScalingLF}. \\
We build on this work and study the impact of scaling on abstractive summarization and what role model asymmetry has in it.
This asymmetry can manifest in various ways, such as the number of layers and hidden units in the encoder and decoder and the type of attention mechanisms used. \\
In this paper, we explore the role of asymmetry in the number of layers in encoder-decoder language modeling for summarization and its impact on the performance of these models. As shown in Figure \ref{fig:speed}, the symmetry of pruning drives the impact on accuracy and inference speedups for sequence-to-sequence models. \\
The following research questions drive our work: 
\begin{itemize}
    \item What scaling laws can be observed in abstractive summarization?
    \item What impact does encoder-decoder asymmetry have on abstractive summarization accuracy? 
    \item What impact does encoder-decoder asymmetry have on abstractive summarization inference efficiency?
    \item What is asymmetries impact on accuracy and inference efficiency does scale have in encoder-decoder models for abstractive summarization? 
\end{itemize}
It is in answering these questions that we deliver the following contributions: 
\begin{itemize}
\item We present the first robust study on scaling laws applied to the compression of sequence-to-sequence modeling. 
\item We demonstrate that the asymmetric inference cost of sequence-to-sequence models leads to asymmetric pruning for optimal inference efficient compression.
\item We empirically demonstrate on a wide variety of benchmarks how Asymmetric Compression can lead to a 2.7x inference speedup with no loss in accuracy on the XSUM dataset.
\end{itemize}
\section{Related Work}
\textbf{Transformer Based Language Models} such as BERT \cite{Devlin2019BERTPO} and T5 \cite{Raffel2020ExploringTL} provide contextual language representations built on the Transformer architecture \cite{Vaswani2017AttentionIA} which can be specialized and adapted for specific tasks and domains \cite{Lee2020BioBERTAP}. Using these models, it becomes relatively easy to excel at a broad range of natural language processing tasks such as question answering, text classification, and sentiment analysis. \\
\begin{table*}[htb!]
    \centering
    \caption{Information about the architecture and attributes of the FLAN-T5 models}
    \scalebox{0.62}{
    \begin{tabular}{|l|l|l|l|l|l|l|l|l|l|}
    \hline
        Model & Size(MBs) & Parameters & Encoder Layers & Parameters Encoder & Decoder Layers & Parameters decoder & Ratio End:Dec & Hidden Size \\ \hline
        Flan-t5-small \footnote{https://huggingface.co/google/flan-t5-small} & 146 & 60511616 & 8 & 35332800 & 8 & 41628352 & 0.849 & 512 \\ \hline
        Flan-t5-base \footnote{https://huggingface.co/google/flan-t5-base} & 472 & 222903552 & 12 & 109628544 & 12 & 137949312 & 0.795 & 768 \\ \hline
        Flan-t5-large \footnote{https://huggingface.co/google/flan-t5-large} & 1500 & 750251008 & 24 & 341231104 & 24 & 441918976 & 0.772 & 1024 \\ \hline
    \end{tabular}}
    \label{tab:models}
\end{table*}
\textbf{Scaling Laws} has become an increasingly important area of study as models' size and training data grows. Performance of the transformer-based language model improves with the relation to model size \cite{Radford2018ImprovingLU} and that larger models outperform smaller models \cite{Brown2020LanguageMA} on most NLP tasks. Increasing the training corpus size can lead to large improvements in performance, and model sizes can have a \textit{optimal} training data size \cite{Hoffmann2022TrainingCL}. Li et al. (2020) \cite{Li2020TrainLT} explore the relationship between model size and training efficiency finding larger models train faster and are more robust to pruning and quantization \cite{Na2022TrainFT}. \\
\textbf{Asymmetrical in sequence-to-sequence models} broadly refers to non-uniformity between encoder and decoder model shape or attributes. Training and inference procedures should match as closely as possible \cite{Ranzato2015SequenceLT} \cite{Mihaylova2019ScheduledSF} as improvements in training loss during optimization result in improvements in model performance during Inference. While this may lead to the best model performance, it ignores the variable inference cost of sequence-to-sequence models.  \\
During Inference, latency is dominated by the asymmetric execution of the language model. The auto-encoding encoder executes once over the entire input sequence, while the auto-regressive decoder executes iteratively until an end-of-sequence token is produced. \\
Kasai et al. demonstrated how the sequence-to-sequence language model performance for machine translation is dominated by the encoder depth \cite{Kasai2020DeepES}. Tay et al. 2021 extend this work by finding a \textit{DeepNarrow} which shows that for broad language modeling, it is possible to have 50\% fewer parameters and a 40\% faster inference with no loss in accuracy \cite{Tay2021ScaleEI}. \\
\textbf{Efficient Inference} for language modeling is a growing area of study that broadly focuses on reducing the inference cost without losses in accuracy. \\
Unstructured Pruning has been broadly studied \cite{Han2015ADN}  \cite{Sanh2020MovementPA} \cite{Kurti2022TheOB} \cite{Zafrir2021PruneOF} \cite{Campos2022SparseBERTSM} but realizing speedups can be difficult. \\ Structured Pruning removes fundamental structural components in a language model such as individual attention heads \cite{Voita2019AnalyzingMS} or entire model layers such as transformer encoders \cite{sanh2019distilbert}. Rosenfeld et al. 2020 demonstrate that unstructured pruning impacts follow scaling laws \cite{Rosenfeld2020OnTP} where larger models can be pruned with greater ease. \\
\textbf{Compressing Sequence-to-sequence} is a growing area of study where approaches from regular, efficient Inference has shown some transfer ability. Shleifer et al. show that it is possible to gain 1.93x speedup on a BART summarization model by applying structural pruning \cite{Shleifer2020PretrainedSD} but find compression approaches differ in their success depending on the dataset. Leveraging semi-structured pruning, Lagunas et al. can gain a 1.19 speedup \cite{Lagunas2021BlockPF} for minor losses in accuracy. While they find that the encoder is easier to prune than the decoder, they do not use this evidence of asymmetry to speed up performance further. \\
Li et al. investigate how to enable quantization, finding that without specialized distillation during quantization, performance collapses \cite{Li2022DQBARTES}.
Leveraging that generation occurs iteratively, and some tokens are easier to generate than other CALM \cite{Schuster2022ConfidentAL} apply early exiting to improve inference speed by 1.4x. While existing work has found interest in asymmetry, it has not been studied directly, nor has relationships in model scale been explored. \\
While there are other approaches such as knowledge distillation \cite{Hinton2015DistillingTK} \cite{sanh2019distilbert} \cite{Jiao2020TinyBERTDB}, quantization \cite{Zafrir2019Q8BERTQ8}, early exiting \cite{Xin2020DeeBERTDE} and token pruning \cite{Kim2021LearnedTP} these are not the focus on our work as understanding the impact of many variables together limits the depth of our exploration. We leave further study of the interplay between summarization and quantization, unstructured pruning, structured pruning, and knowledge distillation for future work. 
\section{Scale and Abstractive Summarization}
\subsection{Background}
\textbf{Sequence-to-sequence language models} such as BART \cite{lewis-etal-2021-paq}, T5 \cite{Raffel2020ExploringTL}, and PEGASUS \cite{Zhang2020PEGASUSPW} combine transformer encoders and decoders to produce models which can adapt to novel tasks and reach top performance on tasks ranging from information retrieval \cite{Nogueira2020DocumentRW} to summarization \cite{Raffel2020ExploringTL}. \\
We focus on the instruction-tuned FLAN-T5 models \cite{Wei2021FinetunedLM} as their performance is competitive and they feature wide variations in model size ranging from 60 million to 11 billion parameters and given the cost of training the larger variants, focus on the small, base, and large variants. Details on model size and architecture can be found in table \ref{tab:models}.\\
\textbf{Abstractive summarization} is a method of sequence compression where a source document $D$ is transformed into a target document $d_{sum}$, which is shorter but faithful to the input. \\
\textbf{Datasets} of use are a combination of public and academic benchmarks and a proprietary web search dataset. The CNN/DailyMail (CNNDM) \cite{see-etal-2017-get} and XSUM \cite{Narayan2018DontGM} datasets are based on the summarization of English new language models. The Query Independent Web Summary (QIWS) is a proprietary corpus of abstractive summaries of web pages that are used to create informative contextual snippets for search engine users. It is important to note the differences in compression factor in each dataset as each impact how decoder-driven inference latency is. Further information on the makeup of each dataset can be found in table \ref{tab:datasets}. \\
\begin{table*}[!ht]
    \centering
    \small
    \caption{Impact of scale on inference throughput for abstractive summarization models trained on the XSUM dataset. Latency is measured in MS/batch and the impact is the impact to latency vs. the small model}
    \begin{tabular}{|l|l|l|l|l|l|l|l|l|}
    \hline
         Model & R-2 & Gain & BS 1 Latency & Impact & BS 8 Latency & Impact & BS 16 Latency & Impact \\ \hline
        small & 17.55 & 0.00\% & 138 & 1 & 230 & 1 & 330 & 1 \\ \hline
        base & 19.77 & 12.63\% & 199 & 1.44 & 550 & 2.39 & 931 & 2.82 \\ \hline
        large & 21.15 & 20.51\% & 445 & 3.22 & 1480 & 6.43 & 2700 & 8.18 \\ \hline
    \end{tabular}
    \label{tab:cnndm-scale-inference}
\end{table*}
\begin{table*}[!ht]
    \centering
    \small
    \caption{Impact of scale on inference throughput for abstractive summarization models trained on the QIWS dataset. Latency is measured in MS/batch and the impact is the impact to latency vs. the small model}
    \begin{tabular}{|l|l|l|l|l|l|l|l|l|}
    \hline
        Model & R-2 & Gain & BS 1 Latency & Impact & BS 8 Latency & Impact & BS 16 Latency & Impact \\ \hline
        small & 29.03 & 0 & 524 & 1 & 653 & 1 & 729 & 1 \\ \hline
        base & 34.19 & 17.77\% & 746 & 1.42 & 1060 & 1.62 & 1310 & 1.80 \\ \hline
        large & 37.37 & 28.72\% & 1,430 & 2.73 & 2240 & 3.43 & 3320 & 4.55 \\ \hline
    \end{tabular}
    \label{tab:qiws-scale-inference}
\end{table*}
\begin{table*}[!ht]
    \centering
    \small
    \caption{Impact of scale on inference throughput for abstractive summarization models trained on the CNNDM dataset.Latency is measured in MS/batch and the impact is the impact to latency vs. the small model}
    \begin{tabular}{|l|l|l|l|l|l|l|l|l|}
    \hline
         Model & R-2 & Gain & BS 1 Latency & Impact & BS 8 Latency & Impact & BS 16 Latency & Impact \\ \hline
        small & 11.09 & 0 & 171 & 1.00 & 252 & 1.00 & 344 & 1.00 \\ \hline
        base & 15.69 & 41.50\% & 255 & 1.49 & 550 & 2.18 & 845 & 2.46 \\ \hline
        large & 16.34 & 47.41\% & 525 & 3.07 & 1370 & 5.44 & 2300 & 6.69 \\ \hline
    \end{tabular}
    \label{tab:xsum-scale-inference}
\end{table*}
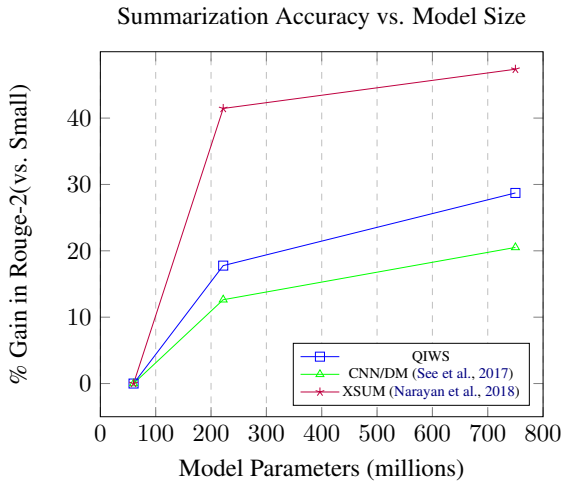
\begin{figure}
    \centering
    \begin{tikzpicture}
\scalebox{0.85}{
\begin{axis}[
    title={Summarization Accuracy vs. Model Size},
    ylabel={\% Gain in Rouge-2(vs. Small)},
    xlabel={Model Parameters (millions)},
    ymin=-5, ymax=50,
    xmin=0, xmax=800,
    ytick={0,10,20,30,40},
    xtick={0,100,200,300,400,500,600,700,800},
    legend pos=south east,
    xmajorgrids=true,
    grid style=dashed,
    legend style={nodes={scale=0.6, transform shape}}, 
    legend image post style={}
]
\addplot[
    color=blue,
    mark=square,
    ]
    coordinates {
    (60,0) ( 222,17.77) (750,28.72) 
    };
\addplot[
    color=green,
    mark=triangle,
    ]
    coordinates {
    (60, 0) (222, 12.63) (750,20.51)
    };  
\addplot[
    color=purple,
    mark=star,
    ]
    coordinates {
    (60,0) (222, 41.45) (750,47.36)
    };
\legend{QIWS, CNN/DM \cite{see-etal-2017-get}, XSUM \cite{Narayan2018DontGM} }
 \end{axis}}
\end{tikzpicture}
    \caption{Model Size vs. Gain to summarization accuracy as measured by the relative Gain in rouge-2 vs. the small model.}
    \label{fig:scale-laws}
\end{figure}
\textbf{Metrics} For each dataset, we evaluate model performance by measuring the ROUGE-1 (R-1), ROUGE-2 (R-2), ROUGE-L (R-L), RougeSum-L (RSL) \footnote{Rouge-L is sentence level vs. RougeSum-L is summary level} \cite{lin-2004-rouge}, and Generation Length (GenL) on the test portion of the dataset. To aid the reproducibility and extension of our work, we experiment using HuggingFace's Transformers \footnote{https://github.com/huggingface/transformers}, release our training and pruning scripts \footnote{https://github.com/spacemanidol/Efficient-Web-Scale-Absractive-Summarization} and model variants for datasets that are publicly available datasets  \footnote{https://huggingface.co/spacemanidol}. 
\subsection{Scaling Laws for Abstract Summarization}
To study the role of scale in abstractive summarization, we train small, base, and large models of the three datasets mentioned above. We do not study the XL (3B) and XXL (11B) as they are expensive and slow to train.\\ For all of our experiments, we train on various hardware but fix the batch size to 64 using gradient accumulation and leverage the hyperparameters in \ref{tab:hyperparams-transfer}. While further hyperparameter optimization and instruction tuning would likely lead to further gains in accuracy, our work is not focused on absolute Gains but on the relative relation of scale.  \\

As shown in \ref{fig:scale-laws}, \ref{tab:scale-qiws}, \ref{tab:scale-cnndm}, and \ref{tab:scale-xsum}, there is a substantial role between scale and performance, but there is a substantial variation across datasets. \\
Datasets with short candidate summaries, such as XSUM, see nearly three times the impact compared to the long summaries of QIWS and CNNDM. During qualitative evaluations, the role of scale can easily be observed as smaller models generate more short keyword summaries while introducing scale makes responses more natural. 
\begin{table}[!ht]
    \centering
    \tiny
    \caption{Relation between scale and asymmetry on model performance on the QIWS dataset. As shown by the results in \textbf{bold} pruning only the decoder leads to less degradation than just the encoder or both, across all scales.  }
    \scalebox{0.9}{
    \begin{tabular}{|l|l|l|l|l|l|l|l|}
    \hline
        \multicolumn{2}{l}{} &  \multicolumn{2}{l}{Small} & \multicolumn{2}{l}{Base} & \multicolumn{2}{l}{Large}  \\ \hline 
        $l_{enc}$ & $l_{dec}$ & R-2 & $R$ & R-2 & $R$ & R-2 & $R$ \\ \hline
        6 & 6 & 29.03 & 100.00\% & 34.19 & 100.00\% & 37.37 & 100.00\% \\ \hline
        6 & 5 & 28.90 & 99.55\% & 34.00 & 99.44\% & 37.59 & 100.59\% \\ \hline
        6 & 4 & 28.56 & 98.40\% & 34.50 & 100.91\% & 36.56 & 97.84\% \\ \hline
        6 & 3 & 27.94 & \textbf{96.24\%} & 33.70 & \textbf{98.58\%} & 35.74 & \textbf{95.64\%} \\ \hline
        6 & 2 & 24.85 & 85.61\% & 31.93 & 93.38\% & 35.13 & 94.01\% \\ \hline
        6 & 1 & 15.41 & 53.08\% & 28.05 & 82.03\% & 33.69 & 90.15\% \\ \hline
        \midrule
        5 & 6 & 27.92 & 96.17\% & 33.57 & 98.18\% & 36.39 & 97.38\% \\ \hline
        4 & 6 & 27.75 & 95.60\% & 33.06 & 96.69\% & 35.90 & 96.07\% \\ \hline
        3 & 6 & 25.20 & \textbf{86.82\%} & 32.23 & \textbf{94.28\%} & 34.22 & \textbf{91.58\%} \\ \hline
        2 & 6 & 23.67 & 81.55\% & 27.47 & 80.35\% & 33.42 & 89.43\% \\ \hline
        1 & 6 & 18.23 & 62.79\% & 25.57 & 74.78\% & 30.31 & 81.11\% \\ \hline
        \midrule
        5 & 5 & 26.82 & 92.38\% & 32.88 & 96.18\% & 36.32 & 97.20\% \\ \hline
        4 & 4 & 26.62 & 91.72\% & 32.81 & 95.96\% & 35.98 & 96.29\% \\ \hline
        3 & 3 & 23.12 & \textbf{79.64\%} & 28.70 & \textbf{83.95\%} & 33.00 & \textbf{88.31\%} \\ \hline
        2 & 2 & 19.14 & 65.92\% & 26.53 & 77.60\% & 30.78 & 82.38\% \\ \hline
        1 & 1 & 6.09 & 20.99\% & 19.64 & 57.43\% & 22.77 & 60.94\% \\ \hline
    \end{tabular}}
    \label{tab:qiws-r2-asym}
\end{table}
\subsection{Inference Benchmark}
To evaluate the impact of asymmetry on inference, we run experiments on the throughput of each model. Using an A10 GPU and the models from our QIWS datasets, we evaluate performance with a max sequence length of 1024, a max summary of 256, and batch sizes 1, 8, and 16 using native inference in PyTorch. We report the mean and standard deviation of timings on seven runs. \\
In comparing the impact of scale on R-2 vs. the effects on latency across batch sizes in \ref{tab:cnndm-scale-inference}, \ref{tab:xsum-scale-inference}, \ref{tab:qiws-scale-inference} it becomes clear that larger models are more expensive to execute significantly as batch sizes increase. This is because of potential differences in output length within a batch as the batch completes when all sequences have produced an \textit{EOS} token. To alleviate this issue bottleneck, improved streaming methods for improved batching have been proposed \cite{Yang2020ASA} but can be challenging to manage. 
\section{To Asymmetry and Beyond}
While prior work has studied how to improve inference and tangentially explored the asymmetry between the encoder and decoder, we study that explicitly and across model scales. We focus our studies on \textbf{structural pruning} as inference gains are easy to realize, and this approach is highly compatible with other methods like quantization and unstructured pruning. We do not study how asymmetry is impacted by unstructured pruning or quantization as these methods are difficult to combine optimized libraries like FasterTransformers\footnote{https://github.com/NVIDIA/FasterTransformer}.  \\
Following Shleifer et al., we adopt the "\textbf{S}hink and \textbf{t}hen \textbf{f}ine" (STF) tune approach for compression. First, a model is trained until convergence on a fine-tuning summarization task. Then, entire layers are removed from the encoder, decoder, or both, and the model is further fine-tuned until it has re-converged. We do not study the use of knowledge distillation to avoid the additional training overhead without guaranteed improvements following Shleifer et al.'s results. \\
Each model we study has a uniform number of encoder and decoder layers, so we prune only the encoders, decoders, and a symmetric combination of the two combinations. We used our three scales of uncompressed models (small, base, large), and we pruned the model in multiples of 1 on the encoder, the decoder, and both. After pruning, models are fine-tuned again and evaluated. This means that for each dataset, we have 16 variants for each model size leading to 48 models per dataset and 144 models overall. \\
Given the wide number of models and the cost of multiple seeds or model-specific optimization, we train each model once and do not optimize the parameters for each model. While this leads to a worse-than-ideal performance, our goal is not to hyper-optimize models but explore where there is high sensitivity. To save space, we use the shorthand $l_{enc}$ and $l_{dec}$ to refer to the number portion of transformer encoder and decoder layers (out of 6), and $R$ refers to the percentage performance recall vs. uncompressed baseline. Detailed results have been moved to the \ref{appendix:asym-full} to save space.\\
\begin{table}[!ht]
    \centering
    \tiny
    \caption{Relation between scale and asymmetry on model performance on the CNNDM dataset. As shown by the results in \textbf{bold} as the model size grows the impact of pruning becomes more muted}
    \scalebox{0.9}{
    \begin{tabular}{|l|l|l|l|l|l|l|l|}
    \hline
        \multicolumn{2}{l}{} &  \multicolumn{2}{l}{Small} & \multicolumn{2}{l}{Base} & \multicolumn{2}{l}{Large}  \\ \hline 
        $l_{enc}$ & $l_{dec}$ & R-2 & $R$ & R-2 & $R$ & R-2 & $R$ \\ \hline
        6 & 6 & 17.55 & 100.00\% & 19.77 & 100.00\% & 21.15 & 100.00\% \\ \hline
        6 & 5 & 17.68 & 100.74\% & 19.92 & 100.76\% & 21.30 & 100.69\% \\ \hline
        6 & 4 & 17.27 & 98.36\% & 19.85 & 100.42\% & 21.32 & 100.81\% \\ \hline
        6 & 3 & 16.40 & \textbf{93.43\%} & 18.85 & \textbf{95.37\%} & 21.08 & \textbf{99.66\%} \\ \hline
        6 & 2 & 15.35 & 87.42\% & 18.68 & 94.51\% & 20.67 & 97.73\% \\ \hline
        6 & 1 & 11.33 & 64.57\% & 16.48 & 83.38\% & 19.49 & 92.12\% \\ \hline
        \midrule
        5 & 6 & 17.69 & 100.81\% & 19.92 & 100.76\% & 21.13 & 99.88\% \\ \hline
        4 & 6 & 17.35 & 98.84\% & 19.67 & 99.50\% & 20.83 & 98.47\% \\ \hline
        3 & 6 & 16.80 & \textbf{95.70\%} & 18.85 & \textbf{95.37\%} & 20.53 & \textbf{97.06\%} \\ \hline
        2 & 6 & 15.54 & 88.51\% & 18.22 & 92.14\% & 19.74 & 93.33\% \\ \hline
        1 & 6 & 13.31 & 75.83\% & 17.06 & 86.27\% & 18.68 & 88.31\% \\ \hline
        \midrule
        5 & 5 & 17.07 & 97.23\% & 19.72 & 99.74\% & 21.23 & 100.34\% \\ \hline
        4 & 4 & 16.20 & 92.28\% & 19.17 & 96.99\% & 20.90 & 98.81\% \\ \hline
        3 & 3 & 14.91 & \textbf{84.95\%} & 17.46 & \textbf{88.29\%} & 20.13 & \textbf{95.16\%} \\ \hline
        2 & 2 & 11.97 & 68.17\% & 15.87 & 80.26\% & 18.47 & 87.30\% \\ \hline
        1 & 1 & 6.05 & 34.45\% & 12.23 & 61.88\% & 15.51 & 73.32\% \\ \hline
    \end{tabular}}
    \label{tab:cnndm-r2-asym}
\end{table}
\begin{table}[!ht]
    \centering
    \tiny
    \caption{Scale and Pruning on XSUM dataset}
    \scalebox{0.9}{
    \begin{tabular}{|l|l|l|l|l|l|l|l|}
    \hline
        \multicolumn{2}{l}{} &  \multicolumn{2}{l}{Small} & \multicolumn{2}{l}{Base} & \multicolumn{2}{l}{Large}  \\ \hline 
        $l_{enc}$ & $l_{dec}$ & R-2 & $R$ & R-2 & $R$ & R-2 & $R$ \\ \hline
        6 & 6 & 11.09 & 100.00\% & 15.69 & 100.00\% & 16.34 & 100.00\% \\ \hline
        6 & 5 & 11.61 & 104.74\% & 15.27 & 97.35\% & 19.80 & 121.16\% \\ \hline
        6 & 4 & 11.43 & 103.12\% & 14.91 & 95.03\% & 19.30 & 118.09\% \\ \hline
        6 & 3 & 11.24 & 101.36\% & 15.40 & 98.17\% & 18.92 & 115.77\% \\ \hline
        6 & 2 & 10.53 & 94.98\% & 15.19 & 96.82\% & 17.96 & 109.93\% \\ \hline
        6 & 1 & 6.03 & 54.42\% & 13.73 & 87.53\% & 16.47 & 100.76\% \\ \hline
        \midrule
        5 & 6 & 11.18 & 100.82\% & 15.92 & 101.47\% & 19.43 & 118.88\% \\ \hline
        4 & 6 & 10.61 & 95.68\% & 14.10 & 89.91\% & 18.33 & 112.16\% \\ \hline
        3 & 6 & 10.11 & 91.16\% & 13.84 & 88.21\% & 16.90 & 103.39\% \\ \hline
        2 & 6 & 8.59 & 77.52\% & 12.10 & 77.12\% & 14.97 & 91.61\% \\ \hline
        1 & 6 & 7.70 & 69.43\% & 10.27 & 65.47\% & 12.52 & 76.63\% \\ \hline
        \midrule
        5 & 5 & 10.73 & 96.76\% & 15.72 & 100.22\% & 19.18 & 117.38\% \\ \hline
        4 & 4 & 10.19 & 91.96\% & 14.30 & 91.15\% & 17.56 & 107.43\% \\ \hline
        3 & 3 & 9.50 & 85.69\% & 12.44 & 79.32\% & 15.89 & 97.21\% \\ \hline
        2 & 2 & 7.31 & 65.91\% & 10.67 & 68.05\% & 12.15 & 74.34\% \\ \hline
        1 & 1 & 4.00 & 36.09\% & 7.74 & 49.35\% & 8.96 & 54.86\% \\ \hline
    \end{tabular}}
    \label{tab:xsum-r2-asym}
\end{table}
\subsection{Scale and Pruning}
Looking at abridged results in \ref{tab:qiws-r2-asym}, \ref{tab:cnndm-r2-asym}, and \ref{tab:xsum-r2-asym}, there is a clear scaling law as smaller models see much larger drops in performance when compressed to the same degree. For example, on the QIWS dataset, compression to $\frac{1}{6}$ of the layers on the encoder and decoder cause an 80\% drop in R-2 on a small model but only 40\% on the larger model. This scale comparison is 65\% to 26\% on CNNDM and 64\% to 45\% on XSUM datasets.\\
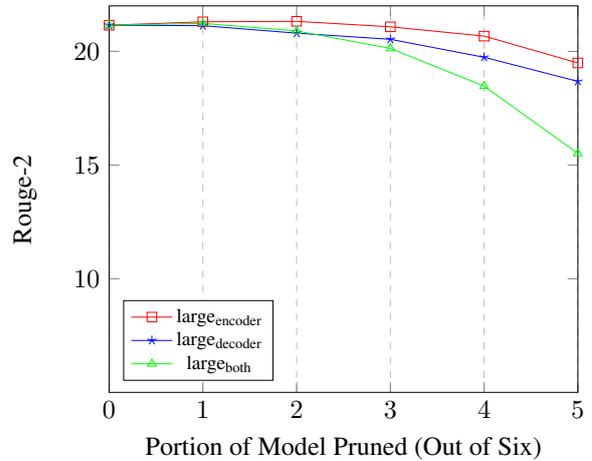
\begin{figure}
    \centering
    \begin{tikzpicture}
\scalebox{0.9}{
\begin{axis}[
    title={Role of scale and compression on CNNDM},
    ylabel={Rouge-2},
    xlabel={Portion of Model Pruned (Out of Six)},
    ymin=5, ymax=22,
    xmin=0, xmax=5,
    ytick={10,15,20},
    xtick={0,1,2,3,4,5},
    legend pos=south west,
    xmajorgrids=true,
    grid style=dashed,
    legend style={nodes={scale=0.75, transform shape}}, 
    legend image post style={}
]
\addplot[
    color=blue,
    mark=square,
    ]
    coordinates {
    (0, 17.55) (1, 17.68) (2, 17.27) (3, 16.4) (4, 15.35) (5,11.33)
    };
\addplot[
    color=green,
    mark=triangle,
    ]
    coordinates {
    (0, 17.55) (1, 17.69) (2, 17.34) (3, 16.8) (4,15.54 ) (5,13.31)
    };  
\addplot[
    color=purple,
    mark=star,
    ]
    coordinates {
    (0, 17.55) (1, 17.07) (2, 16.2) (3, 14.91) (4, 11.97) (5,6.05)
    };

\legend{small\textsubscript{encoder}, small\textsubscript{decoder}, small\textsubscript{both}, }
 \end{axis}}
\end{tikzpicture}
\begin{tikzpicture}
\scalebox{0.9}{
\begin{axis}[
    title={Role of scale and compression on CNNDM},
    ylabel={Rouge-2},
    xlabel={Portion of Model Pruned (Out of Six)},
    ymin=5, ymax=22,
    xmin=0, xmax=5,
    ytick={10,15,20},
    xtick={0,1,2,3,4,5},
    legend pos=south west,
    xmajorgrids=true,
    grid style=dashed,
    legend style={nodes={scale=0.75, transform shape}}, 
    legend image post style={}
]
\addplot[
    color=blue,
    mark=square,
    ]
    coordinates {
    (0, 19.77) (1, 19.92) (2, 19.85) (3, 18.85) (4, 18.68) (5,16.48)
    };
\addplot[
    color=red,
    mark=star,
    ]
    coordinates {
    (0, 19.77) (1, 19.92) (2, 19.67) (3, 18.85) (4,18.22 ) (5,17.06)
    };  
\addplot[
    color=green,
    mark=triangle,
    ]
    coordinates {
    (0, 19.77) (1, 19.72) (2, 19.17) (3, 17.46) (4, 15.87) (5,12.23)
    };

\legend{ base\textsubscript{encoder}, base\textsubscript{decoder}, base\textsubscript{both} }
 \end{axis}}
\end{tikzpicture}
\begin{tikzpicture}
\scalebox{0.9}{
\begin{axis}[
    title={Role of scale and compression on CNNDM},
    ylabel={Rouge-2},
    xlabel={Portion of Model Pruned (Out of Six)},
    ymin=5, ymax=22,
    xmin=0, xmax=5,
    ytick={10,15,20},
    xtick={0,1,2,3,4,5},
    legend pos=south west,
    xmajorgrids=true,
    grid style=dashed,
    legend style={nodes={scale=0.75}}, 
    legend image post style={}
]
\addplot[
    color=red,
    mark=square,
    ]
    coordinates {
    (0, 21.15) (1, 21.30) (2, 21.32) (3, 21.08) (4, 20.67) (5,19.49)
    };
\addplot[
    color=blue,
    mark=star,
    ]
    coordinates {
    (0, 21.15) (1, 21.13) (2,20.8 ) (3, 20.53) (4,19.74 ) (5,18.68)
    };  
\addplot[
    color=green,
    mark=triangle,
    ]
    coordinates {
    (0, 21.15) (1, 21.23) (2, 20.9) (3, 20.13) (4, 18.47) (5,15.51)
    };
\legend{large\textsubscript{encoder}, large\textsubscript{decoder}, large\textsubscript{both} }
 \end{axis}}
\end{tikzpicture}
    \caption{Relationship between model compression, model size, and summarization accuracy measured by rouge-2 vs. Number Layers. small\textsubscript{encoder} refers to a FLAN-T5 small which has only pruned the encoder, small\textsubscript{decoder} for only the decoder, and small\textsubscript{both} for the encoder and decoder}
    \label{fig:scale-laws-pruning}
\end{figure}
Similar scaling results hold with encoder or decoder pruning, where compressing large models lead to a 5x lower loss in performance than small models. As the model's size grows, the impact of decoder vs. encoder-only pruning becomes more muted. On the CNNDM dataset, the gap between the decoder only and encoder only pruned to $\frac{1}{6}$ is 10\% with the FLAN-T5 small but only 4\% with the large variant. When comparing asymmetric and symmetric, the gap is even further pronounced where the small gap is 30\% while the large is 20\%. \\
As shown in Figure \ref{fig:scale-laws-pruning}, the impact of compression becomes more muted as the model size grows. In other words, larger models are more compressible and amenable to asymmetry in this compression. \\
The impact of asymmetry is easiest to understand as it is not surprising that complete pruning of a model leads to higher losses than partial pruning across datasets and model sizes. While this finding is not immediately surprising, evaluating the inference costs becomes important. \\
\subsection{Inference Benchmarks}
We evaluate the impact of asymmetry in a similar method to our scale experiments. Using an A10 GPU, we evaluate performance for summarization on a portion of each model's respective evaluation datasets with a max sequence length of 1024, a max summary length of 256, and batch sizes 1, 8, and 16. We choose these batch sizes to represent streaming workloads (batch size 1), real-time results for the top results from a search query (batch size 8i), and max throughput given the A10's memory budget (batch size 16)\\
\begin{table}[!htb!]
    \centering
    \tiny
    \scalebox{0.8}{
    \begin{tabular}{|l|l|l|l|l|l|l|l|}
    \hline
        \multicolumn{2}{l}{} &  \multicolumn{2}{l}{QIWS} & \multicolumn{2}{l}{CNN/DailyMail} & \multicolumn{2}{l}{XSUM}  \\ \hline 
        \midrule
        $l_{enc}$ & $l_{dec}$ & Impact & Speedup & Impact & Speedup & Impact & Speedup \\ \hline
        6 & 3 & -4.36\% & 1.80 & -0.34\% & 1.65 & 15.77\% & 1.64 \\ \hline
        6 & 2 & -5.99\% & 2.44 & -2.27\% & 2.03 & 9.93\% & 2.07 \\ \hline
        6 & 1 & -9.85\% & 3.83 & -7.88\% & 2.70 & \textbf{0.76\%} & \textbf{2.71} \\ \hline
        \midrule
        3 & 6 & -8.42\% & 1.04 & -2.94\% & 1.14 & 3.39\% & 1.16 \\ \hline
        2 & 6 & -10.57\% & 1.04 & -6.67\% & 1.19 & -8.39\% & 1.21 \\ \hline
        1 & 6 & -18.89\% & 1.06 & -11.69\% & 1.27 & \textbf{-23.37}\% & \textbf{1.30} \\ \hline
        \midrule
        3 & 3 & -11.69\% & 1.91 & -4.84\% & 1.94 & -2.79\% & 2.06 \\ \hline
        2 & 2 & -17.62\% & 2.20 & -12.70\% & 2.78 & -25.66\% & 2.83 \\ \hline
        1 & 1 & -39.06\% & 2.44 & -26.68\% & 4.96 & \textbf{-45.14\%} & \textbf{4.84} \\ \hline
    \end{tabular}}
    \caption{Relationship between accuracy and speedup of encoder only, the decoder only, encoder and decoder pruning on FLAN-T5 Large models on CNN/DM, XSUM, and QIWS. Speedup is measured by comparing the improvements in latency for batch size one vs. the uncompressed baseline. The impact is the relative loss of Rouge-2 of compressed models vs. the uncompressed baseline.}
    \label{tab:asymetry-vs-speedup}
\end{table}\\
Looking at the focused set of results for large models across datasets in table \ref{tab:asymetry-vs-speedup} on the impact of R-2 vs. inference speedup, we can see a clear relationship between asymmetry and inference efficiency. While detailed inference results can be found in the appendix \ref{sec:inference-benchmarks} on this focused set of results, we can see that pruning only the encoder leads to no more than 30\% improvement in inference efficiency at a sizable loss in accuracy. Pruning the model symmetrically leads to realizable inference improvements of up to ~5x at the expense of summarization accuracy. \\ 
Alternatively, when only the decoder is pruned, it is possible to see most of the inference speedups seen during constant pruning with a substantially lower impact on accuracy. On the CNN/DM dataset, constant pruning leads to 8\% better inference but losses nearly four times the performance of non-uniform compression.  \\
\begin{table}[!htb!]
    \centering
    \small
    \scalebox{0.7}{
    \begin{tabular}{|l|l|l|l|l|l|l|l|}
    \hline
        \multicolumn{2}{l}{} &  \multicolumn{2}{l}{Small} & \multicolumn{2}{l}{Base} & \multicolumn{2}{l}{Large}  \\ \hline 
        \midrule
        $l_{enc}$ & $l_{dec}$ & Impact & Speedup & Impact & Speedup & Impact & Speedup \\ \hline
        6 & 3 & -3.76\% & 1.79 & -1.42\% & 1.76 & -4.36\% & 1.80 \\ \hline
        6 & 2 & -14.39\% & 2.69 & -6.62\% & 2.13 & -5.99\% & 2.44 \\ \hline
        6 & 1 & -46.92\% & 3.97 & -17.97\% & 3.69 & \textbf{-9.85}\% & \textbf{3.83 }\\ \hline
        \midrule
        3 & 6 & -13.18\% & 1.02 & -5.72\% & 1.04 & -8.42\% & 1.04 \\ \hline
        2 & 6 & -18.45\% & 1.02 & -19.65\% & 1.05 & -10.57\% & 1.04 \\ \hline
        1 & 6 & -37.21\% & 1.03 & -25.22\% & 1.06 & \textbf{-18.89\%} & \textbf{1.06} \\ \hline
        \midrule
        3 & 3 & -20.36\% & 1.40 & -16.05\% & 1.86 & -11.69\% & 1.91 \\ \hline
        2 & 2 & -34.08\% & 1.30 & -22.40\% & 2.48 & -17.62\% & 2.20 \\ \hline
        1 & 1 & -79.01\% & 3.91 & -42.57\% & 3.95 & \textbf{-39.06\%} & \textbf{2.44} \\ \hline
    \end{tabular}}
    \caption{Relationship between accuracy and speedup of encoder only, decoder only, encoder and decoder pruning on FLAN-T5 models on QIWS concerning model size. Speedup is measured by comparing the improvements in latency for batch size one vs. the uncompressed baseline. The impact is the relative loss of Rouge-2 of compressed models vs. the uncompressed baseline.}
    \label{tab:scale-asym-speedup}
\end{table}
\begin{table}[!htb!]
    \centering
    \small
    \scalebox{0.7}{
    \begin{tabular}{|l|l|l|l|l|l|}
    \hline
        $l_{enc}$ & $l_{dec}$ & Impact & Speedup (BS1)&Speedup (BS8) & Speedup (BS16) \\ \hline
        6 & 3 & -0.34\% & 1.65 & 1.18 & 1.15 \\ \hline
        6 & 2 & -2.27\% & 2.03 & 1.25 & 1.22 \\ \hline
        6 & 1 & -7.88\% & \textbf{2.70} & \textbf{1.36} & \textbf{1.29} \\ \hline
        \midrule
        3 & 6 & -2.94\% & 1.14 & 1.48 & 1.54 \\ \hline
        2 & 6 & -6.67\% & 1.19 & 1.68 & 1.89 \\ \hline
        1 & 6 & -11.69\% & \textbf{1.27} & \textbf{2.21} & \textbf{2.43} \\ \hline
        \midrule
        3 & 3 & -4.84\% & 1.94 & 1.96 & 1.97 \\ \hline
        2 & 2 & -12.70\% & 2.78 & 2.88 & 2.92 \\ \hline
        1 & 1 & -26.68\% & \textbf{4.96} & \textbf{5.54} & \textbf{5.64} \\ \hline
    \end{tabular}}
    \caption{Relationship between accuracy and speedup of encoder only, decoder only, encoder and decoder pruning on FLAN-T5 large models on CNN with variation in inference batch size. Speedup is measured by comparing the improvements in latency vs. the uncompressed baseline at various batch sizes. The impact is the relative loss of Rouge-2 of compressed models vs. the uncompressed baseline.}
    \label{tab:scale-bs-speedup}
\end{table}
\section{Discussion}
\subsection{Scale, Inference, and Pruning}
As shown in table \ref{tab:scale-asym-speedup}, the gains found by pruning are extremely consistent independently with scaling. Pruning only the encoder leads to a 4-6\% improvement in latency, and pruning just the decoder leads to ~400\%, as does uniform compression. This is expected as structural pruning removes a constant portion of the network, which leads to consistent latency gains irrespective of model scale. 
\begin{figure}
    \centering
    \begin{tikzpicture}
\scalebox{0.8}{
\begin{axis}[
    title={Genl vs. Model Pruning on CNNDM},
    ylabel={Generation Length (tokens)},
    ymin=55, ymax=80,
    ytick={60,65,70,75,80},
    xlabel={Portion of Model Pruned (Out of Six)},
    xmin=0, xmax=5,
    xtick={0,1,2,3,4,5},
    legend pos=south west,
    xmajorgrids=true,
    grid style=dashed,
    legend style={nodes={scale=0.7, transform shape}}, 
    legend image post style={}
]
\addplot[
    color=blue,
    mark=square,
    ]
    coordinates {
    (0, 77.62) (1, 76.46) (2, 78.63) (3, 75.69) (4, 75.08) (5,67.99)
    };
\addplot[
    color=green,
    mark=triangle,
    ]
    coordinates {
    (0,  77.62) (1,77.81 ) (2, 76.22) (3, 78.13) (4, 77.77) (5,70.79)
    };  
\addplot[
    color=purple,
    mark=heart,
    ]
    coordinates {
    (0,  77.62) (1, 77.93) (2, 79.83 ) (3, 74.61 ) (4, 78.53) (5, 60.03)
    };
\addplot[
    color=orange,
    mark=diamond,
    ]
    coordinates {
    (0,  71.86) (1, 74.38 ) (2, 70.74 ) (3,74.76 ) (4, 67.52 ) (5, 67.67)
    };
\addplot[
    color=yellow,
    mark=ball,
    ]
    coordinates {
    (0,  71.86) (1,74.38 ) (2, 74.34) (3, 74.76) (4, 77.04) (5, 73.36)
    };  
\addplot[
    color=brown,
    mark=o,
    ]
    coordinates {
    (0,  71.86) (1, 73.56 ) (2, 74.59) (3, 72.27) (4, 69.08) (5, 66.70)
    };
\addplot[
    color=pink,
    mark=x,
    ]
    coordinates {
    (0,  71.01) (1, 70.90 ) (2, 71.85 ) (3, 72.81 ) (4, 73.39) (5,68.58)
    };
\addplot[
    color=black,
    mark=y,
    ]
    coordinates {
    (0,  71.01) (1,71.04 ) (2, 71.59) (3, 73.28) (4, 73.47) (5,76.05)
    };  
\addplot[
    color=purple,
    mark=star,
    ]
    coordinates {
    (0,  71.01) (1, 70.9 ) (2, 71.85 ) (3, 72.81 ) (4, 73.39 ) (5, 68.58)
    };
\legend{small\textsubscript{encoder}, small\textsubscript{decoder}, small\textsubscript{both}, base\textsubscript{encoder}, base\textsubscript{decoder}, base\textsubscript{both},large\textsubscript{encoder}, large\textsubscript{decoder}, large\textsubscript{both} }
 \end{axis}}
\end{tikzpicture}
    \caption{Role of scale and compression on generation length}
    \label{fig:scale-genlen1}
\end{figure}
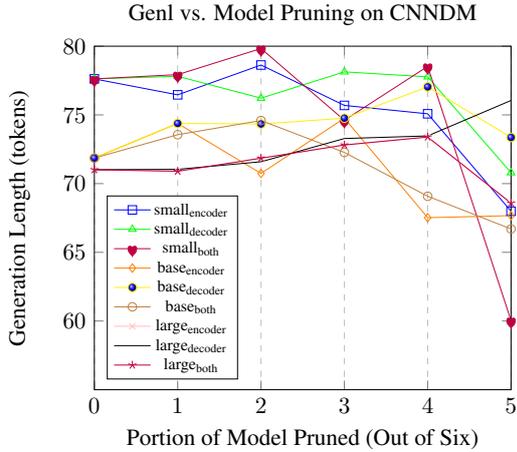
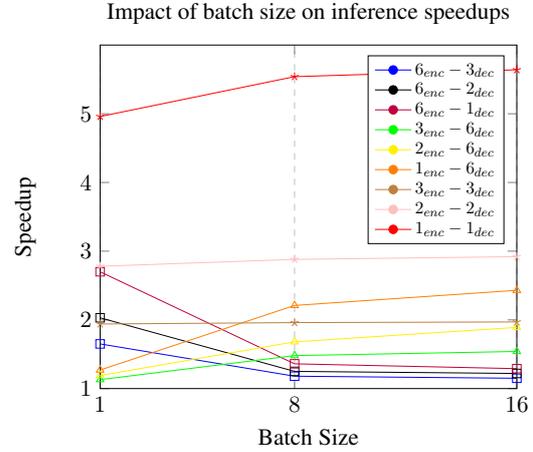
\begin{figure}
    \centering
    \begin{tikzpicture}
\scalebox{0.8}{
\begin{axis}[
    title={Impact of batch size on inference speedups},
    ylabel={Speedup},
    xlabel={Batch Size},
    ymin=1, ymax=6,
    xmin=1, xmax=16,
    ytick={1,2,3,4,5},
    xtick={1, 8, 16},
    legend pos=north east,
    xmajorgrids=true,
    grid style=dashed,
    legend style={nodes={scale=0.74, transform shape}}, 
    legend image post style={mark=*}
] 
\addplot[
    color=blue,
    mark=square,
    ]
    coordinates {
    (1, 1.65) (8, 1.18) (16,1.15)
    };
\addplot[
    color=black,
    mark=square,
    ]
    coordinates {
    (1, 2.03) (8, 1.25) (16,1.22)
    };
\addplot[
    color=purple,
    mark=square,
    ]
    coordinates {
    (1, 2.70) (8, 1.36) (16,1.29)
    };
\addplot[
    color=green,
    mark=triangle,
    ]
    coordinates {
    (1, 1.13) (8, 1.48) (16,1.54)
    };
\addplot[
    color=yellow,
    mark=triangle,
    ]
    coordinates {
    (1, 1.19) (8, 1.68) (16,1.89)
    };
\addplot[
    color=orange,
    mark=triangle,
    ]
    coordinates {
    (1, 1.27) (8, 2.21) (16,2.43)
    };
\addplot[
    color=brown,
    mark=star,
    ]
    coordinates {
    (1, 1.94) (8, 1.96) (16,1.97)
    }; 
\addplot[
    color=pink,
    mark=star,
    ]
    coordinates {
    (1, 2.78) (8, 2.88) (16,2.92)
    }; 
\addplot[
    color=red,
    mark=star,
    ]
    coordinates {
    (1, 4.96) (8, 5.54) (16,5.64)
    };  
\legend{$6_{enc}-3_{dec}$, $6_{enc}-2_{dec}$,$6_{enc}-1_{dec}$,$3_{enc}-6_{dec}$,$2_{enc}-6_{dec}$,$1_{enc}-6_{dec}$,$3_{enc}-3_{dec}$ ,$2_{enc}-2_{dec}$ ,$1_{enc}-1_{dec}$ }
 \end{axis}}
\end{tikzpicture}
    \caption{Relationship between inference batch size and realized inference speedup with uniform and no uniform pruning of FLAN-T5 large on CNNDM}
    \label{fig:batch-size-scale}
\end{figure}
\subsection{Scale, Pruning and Generated length}
\begin{figure}
    \centering
    \begin{tikzpicture}
\scalebox{0.8}{
\begin{axis}[
    title={Genl vs. Model Size},
    ylabel={Generation Length (tokens)},
    xlabel={Model Parameters (millions)},
    ymin=20, ymax=100,
    xmin=0, xmax=800,
    ytick={20,30,40,50,60,70},
    xtick={0,100,200,300,400,500,600,700,800},
    legend pos=north east,
    xmajorgrids=true,
    grid style=dashed,
    legend style={nodes={scale=1.0, transform shape}}, 
    legend image post style={}
]
\addplot[
    color=blue,
    mark=heart,
    ]
    coordinates {
    (60,62.79) ( 222,62.91) (750,62.85) 
    };
\addplot[
    color=green,
    mark=square,
    ]
    coordinates {
    (60, 77.62) (222, 71.86) (750,71.01)
    };  
\addplot[
    color=purple,
    mark=triangle,
    ]
    coordinates {
    (60,28.01) (222, 25.92) (750,26.74)
    };
\legend{QIWS, CNN/DM , XSUM}
 \end{axis}}
\end{tikzpicture}
  \caption{Role of scale on generation length}
    \label{fig:scale-genlen}
\end{figure}
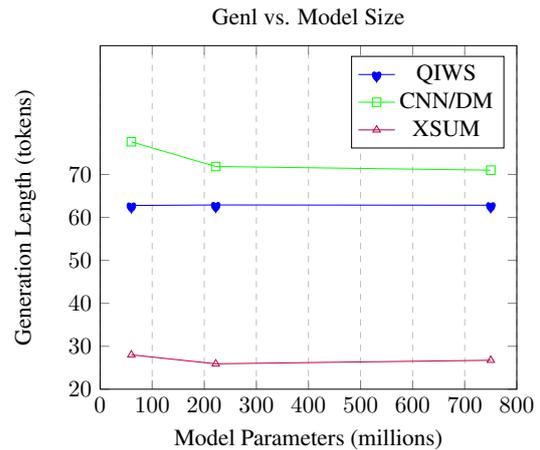
Despite expecting a significant trend in the role of scale and pruning in a generation, we do not see any noticeable trends. As shown in figures \ref{fig:scale-genlen} and \ref{fig:scale-genlen1}, there is no discernible trend of the Role of scale and pruning in generation length. There is a minor jump in generation length from FLAN-T5 small to FLAN-T5 base across all datasets but no such jump from FLAN-T5 base to FLAN-T5 large. We believe this is because the smaller models are less fluent and need more tokens to ensure accurate coverage. As models scale, this is no longer needed, and the models converge to a uniform summary length. 
\subsection{Asymmetry with large batches}
Despite the allures of asymmetrical pruning, it is not without fault. As shown in table \ref{tab:scale-bs-speedup} and Figure \ref{fig:batch-size-scale}, the improvements in inference efficiency are heavily influenced by the batch size. When the batch size is minimal, the difference in the type of non-uniformity has a significant impact on inference efficiency. As batches scale, the speedup from encoder only or decoder only becomes much closer and becomes minor when compared to uniform methods. This indicates why further work on improving generative inference methods is highly relevant, as this problem impacts other efficiency-driven processes like CALM \cite{Schuster2022ConfidentAL}.
\section{Conclusion and Future Work}
In this work, we explore the role of symmetry in the pruning of sequence-to-sequence models for abstractive summarization, finding that pruning asymmetrically can lead to inference speedups with low losses in accuracy. Our work also explores the relationship between model scale and the sensitivity to pruning, finding that larger models see lower losses when pruned. This compresses FLAN-T5 models to deliver ~3x inference gains with a ~1 Rouge-2 point loss. \\
In future work, we seek to study how pseudo labeling, early exiting, and quantization can be combined to improve further the inference efficiency of sequence-to-sequence models.  
\bibliography{anthology,custom}
\bibliographystyle{acl_natbib}
\appendix
\section{Appendix}
\label{sec:appendix}
\subsection{Training Details}
\label{sec:training}
In all of our experiments, we leverage the parameters shown in \ref{tab:hyperparams-transfer} on the datasets shown in \ref{tab:datasets}
\begin{table*}[!htb]
    \centering
    \caption{Statistics for the abstractive summarization datasets which we study. Source and Summary refer to the number of words in each, and the compression factor is the ratio between the two on the train portion of the dataset.}
    \begin{tabular}{|l|l|l|l|l|l|l|}
    \hline
        Dataset  & Train  & Validation & Test & Source & Summary & Compression \\ \hline
        CNNDM \footnote{https://huggingface.co/datasets/cnn\_dailymail} & 287,113 & 13,368 & 11,490 & 691.87 & 51.57 & 14.80 \\ \hline
        XSUM \footnote{https://huggingface.co/datasets/xsum}& 204,045 & 11,332 & 11,334 & 373.86 & 21.09 & 18.70\\ \hline
        QIWS & 10000 & 1000 & 1000 & 1410.12 & 73.78 & 19.11 \\ \hline
    \end{tabular}
    \label{tab:datasets}
\end{table*}
\begin{table}
        {
            \begin{tabular}{l|c}
            \toprule
           HyperParameter & Value\\
            \midrule
            Training Length & 3,10 Epochs \\
            \midrule
            Initial learning rate & 1e-4\\
            Learning rate schedule &  constant \\
            \midrule
                Batch size & 64 \\
            \midrule
                Weight Decay & 0.01, 0.05, 0.1 \\
            \midrule
            \bottomrule
            \end{tabular}
        }
    \caption{Training Hyperparameters for summarization experiments}
    \label{tab:hyperparams-transfer}
\end{table}
\subsection{Scale and Abstractive summarization}
The role of model scale on performance on the QIWS, CNN/DM, and XSUM datasets can be found in tables \ref{tab:scale-cnndm},\ref{tab:scale-qiws}, and \ref{tab:scale-xsum}
\begin{table*}[!htb!]
    \centering
    \small
    \scalebox{0.98}{
    \begin{tabular}{|l|l|l|l|l|l|l|l|l|l|l|}
    \hline
        Model & R-1 & Impact & R-2 & Impact & RSL & Impact & R-L & Impact & Genl & Impact \\ \hline
        small & 50.22 & 0.00\% & 29.03 & 0.00\% & 45.87 & 0.00\% & 40.19 & 0.00\% & 62.79 & 0.00\% \\ \hline
         base&  54.84 & 9.20\% & 34.19 & 17.77\% & 50.38 & 9.83\% & 44.68 & 11.18\% & 62.91 & 0.19\% \\ \hline
         large &57.81 & 15.11\% & 37.37 & 28.72\% & 53.14 & 15.84\% & 48.16 & 19.84\% & 62.85 & 0.10\% \\ \hline
    \end{tabular}}
    \caption{Impact of Scale on summarization performance on QIWS dataset}
    \label{tab:scale-qiws}
\end{table*}
\begin{table*}[!htb!]
    \centering
    \small
    \scalebox{0.98}{
    \begin{tabular}{|l|l|l|l|l|l|l|l|l|l|l|}
    \hline
        Model & R-1 & Impact & R-2 & Impact & RSL & Impact & R-L & Impact & Genl & Impact \\ \hline
        small & 39.31 & 0.00\% & 17.55 & 0.00\% & 36.50 & 0.00\% & 27.97 & 0.00\% & 77.62 & 0.00\% \\ \hline
        base & 42.14 & 7.20\% & 19.77 & 12.63\% & 39.32 & 7.75\% & 30.15 & 7.80\% & 71.86 & -7.42\% \\ \hline
        large & 43.99 & 11.90\% & 21.15 & 20.51\% & 41.12 & 12.68\% & 31.64 & 13.11\% & 71.01 & -8.51\% \\ \hline
    \end{tabular}}
    \caption{Impact of Scale on summarization performance on CNNDM dataset}
    \label{tab:scale-cnndm}
\end{table*}\\
\begin{table*}[!htb]
    \centering
    \small
    \scalebox{0.98}{
    \begin{tabular}{|l|l|l|l|l|l|l|l|l|l|l|}
    \hline
        Model & R-1 & Impact & R-2 & Impact & RSL & Impact & R-L & Impact & Genl & Impact \\ \hline
        small & 33.2675 & 0.00\% & 11.09 & 0.00\% & 26.17 & 0.00\% & 26.17 & 0.00\% & 28.01 & 0.00\% \\ \hline
        base & 38.7782 & 16.56\% & 15.69 & 41.45\% & 31.14 & 19.01\% & 31.15 & 19.04\% & 25.92 & -7.48\% \\ \hline
        large & 39.7125 & 19.36\% & 16.34 & 47.36\% & 31.72 & 21.21\% & 31.72 & 21.23\% & 26.74 & -4.54\% \\ \hline
    \end{tabular}}
    \caption{Impact of Scale on summarization performance on XSUM dataset}
    \label{tab:scale-xsum}
\end{table*}
\subsection{Asymmetry in Summarization}
\label{appendix:asym-full}
The role of the model scale, structural pruning, and asymmetry on performance on the QIWS, CNN/DM, and XSUM datasets can be found in tables \ref{tab:asym-small-xsum},\ref{tab:asym-base-xsum},\ref{tab:asym-large-xsum},\ref{tab:asym-small-cnndm},\ref{tab:asym-base-cnndm},\ref{tab:asym-large-cnndm},\ref{tab:asym-small-qiws},\ref{tab:asym-base-qiws}, and \ref{tab:asym-large-qiws}.
\begin{table*}[!ht]
    \centering
    \caption{The relation between pruning asymmetry and symmetry for a FLAN-T5 small model on the CNN/DailyMail Abstractive Summarization Dataset}
    \small
    \begin{tabular}{|l|l|l|l|l|l|l|l|l|l|l|l|}
    \hline
        $l_{enc}$ & $l_{dec}$ & R-1 & Impact & R-2 & Impact & RSL & Impact & R-L & Impact & GenL & Impact \\ \hline
        8 & 8 & 39.31 & 0.00\% & 17.55 & 0.00\% & 36.50 & 0.00\% & 27.97 & 0.00\% & 77.62 & 0.00\% \\ \hline
        8 & 6 & 39.33 & 0.04\% & 17.68 & 0.74\% & 36.54 & 0.13\% & 28.21 & 0.87\% & 76.46 & -1.49\% \\ \hline
        8 & 5 & 38.75 & -1.42\% & 17.27 & -1.64\% & 36.01 & -1.32\% & 27.91 & -0.23\% & 78.63 & 1.31\% \\ \hline
        8 & 4 & 37.18 & -5.42\% & 16.40 & -6.57\% & 34.46 & -5.58\% & 27.22 & -2.70\% & 75.69 & -2.48\% \\ \hline
        8 & 2 & 35.47 & -9.76\% & 15.35 & -12.58\% & 32.78 & -10.17\% & 26.28 & -6.06\% & 75.08 & -3.27\% \\ \hline
        8 & 1 & 29.27 & -25.55\% & 11.33 & -35.43\% & 26.97 & -26.09\% & 22.33 & -20.18\% & 67.99 & -12.40\% \\ \hline
        6 & 8 & 39.59 & 0.71\% & 17.69 & 0.81\% & 36.80 & 0.83\% & 28.08 & 0.39\% & 77.81 & 0.25\% \\ \hline
        5 & 8 & 39.12 & -0.47\% & 17.35 & -1.16\% & 36.38 & -0.31\% & 27.73 & -0.88\% & 76.22 & -1.80\% \\ \hline
        4 & 8 & 38.57 & -1.87\% & 16.80 & -4.30\% & 35.79 & -1.92\% & 27.15 & -2.92\% & 78.13 & 0.67\% \\ \hline
        2 & 8 & 36.82 & -6.32\% & 15.54 & -11.49\% & 34.00 & -6.84\% & 25.79 & -7.78\% & 77.77 & 0.20\% \\ \hline
        1 & 8 & 33.58 & -14.58\% & 13.31 & -24.17\% & 30.96 & -15.16\% & 23.72 & -15.19\% & 70.79 & -8.79\% \\ \hline
        6 & 6 & 38.59 & -1.82\% & 17.07 & -2.77\% & 35.80 & -1.91\% & 27.55 & -1.52\% & 77.93 & 0.41\% \\ \hline
        5 & 5 & 37.31 & -5.08\% & 16.20 & -7.72\% & 34.60 & -5.19\% & 26.83 & -4.07\% & 79.83 & 2.85\% \\ \hline
        4 & 4 & 35.28 & -10.25\% & 14.91 & -15.05\% & 32.54 & -10.85\% & 25.74 & -7.98\% & 74.61 & -3.88\% \\ \hline
        2 & 2 & 30.79 & -21.66\% & 11.97 & -31.83\% & 28.03 & -23.19\% & 22.88 & -18.19\% & 78.53 & 1.18\% \\ \hline
        1 & 1 & 21.30 & -45.80\% & 6.05 & -65.55\% & 19.57 & -46.39\% & 16.62 & -40.56\% & 60.03 & -22.66\% \\ \hline
    \end{tabular}
    \label{tab:asym-small-cnndm}
\end{table*}

\begin{table*}[!ht]
    \centering
    \caption{The relation between pruning asymmetry and symmetry for a FLAN-T5 base model on the CNN/DailyMail Abstractive Summarization Dataset}
    \small
    \begin{tabular}{|l|l|l|l|l|l|l|l|l|l|l|l|}
    \hline
        $l_{enc}$ & $l_{dec}$ & R-1 & Impact & R-2 & Impact & RSL & Impact & R-L & Impact & GenL & Impact \\ \hline
        12 & 12 & 42.14 & 0.00\% & 19.77 & 0.00\% & 39.32 & 0.00\% & 30.15 & 0.00\% & 71.86 & 0.00\% \\ \hline
        12 & 10 & 42.49 & 0.84\% & 19.92 & 0.76\% & 39.62 & 0.75\% & 30.27 & 0.40\% & 74.38 & 3.51\% \\ \hline
        12 & 8 & 42.28 & 0.34\% & 19.85 & 0.42\% & 39.48 & 0.41\% & 30.35 & 0.64\% & 70.74 & -1.56\% \\ \hline
        12 & 6 & 41.30 & -1.99\% & 18.85 & -4.63\% & 38.44 & -2.25\% & 29.16 & -3.28\% & 74.76 & 4.04\% \\ \hline
        12 & 4 & 40.31 & -4.34\% & 18.68 & -5.49\% & 37.71 & -4.10\% & 29.45 & -2.33\% & 67.52 & -6.04\% \\ \hline
        12 & 2 & 36.75 & -12.80\% & 16.48 & -16.62\% & 34.22 & -12.97\% & 27.61 & -8.43\% & 67.67 & -5.82\% \\ \hline
        10 & 12 & 42.49 & 0.84\% & 19.92 & 0.76\% & 39.62 & 0.75\% & 30.27 & 0.40\% & 74.38 & 3.51\% \\ \hline
        8 & 12 & 42.27 & 0.31\% & 19.67 & -0.50\% & 39.41 & 0.22\% & 29.99 & -0.52\% & 74.34 & 3.45\% \\ \hline
        6 & 12 & 41.30 & -1.99\% & 18.85 & -4.63\% & 38.44 & -2.25\% & 29.16 & -3.28\% & 74.76 & 4.04\% \\ \hline
        4 & 12 & 40.51 & -3.86\% & 18.22 & -7.86\% & 37.66 & -4.23\% & 28.42 & -5.75\% & 77.04 & 7.21\% \\ \hline
        2 & 12 & 39.03 & -7.38\% & 17.06 & -13.73\% & 36.15 & -8.08\% & 27.23 & -9.69\% & 73.36 & 2.09\% \\ \hline
        10 & 10 & 42.19 & 0.13\% & 19.72 & -0.26\% & 39.38 & 0.14\% & 30.12 & -0.11\% & 73.56 & 2.37\% \\ \hline
        8 & 8 & 41.64 & -1.18\% & 19.17 & -3.01\% & 38.83 & -1.26\% & 29.60 & -1.84\% & 74.59 & 3.80\% \\ \hline
        6 & 6 & 39.33 & -6.67\% & 17.46 & -11.71\% & 36.67 & -6.74\% & 28.07 & -6.92\% & 72.27 & 0.57\% \\ \hline
        4 & 4 & 36.99 & -12.23\% & 15.87 & -19.74\% & 34.43 & -12.43\% & 26.63 & -11.68\% & 69.08 & -3.87\% \\ \hline
        2 & 2 & 30.99 & -26.45\% & 12.23 & -38.12\% & 28.43 & -27.71\% & 23.28 & -22.79\% & 66.70 & -7.18\% \\ \hline
    \end{tabular}
    \label{tab:asym-base-cnndm}
\end{table*}
\begin{table*}[!ht]
    \centering
    \caption{The relation between pruning asymmetry and symmetry for a FLAN-T5 large model on the CNN/DailyMail Abstractive Summarization Dataset}
    \small
    \begin{tabular}{|l|l|l|l|l|l|l|l|l|l|l|l|}
    \hline
        $l_{enc}$ & $l_{dec}$ & R-1 & Impact & R-2 & Impact & RSL & Impact & R-L & Impact & GenL & Impact \\ \hline
        24 & 24 & 43.99 & 0.00\% & 21.15 & 0.00\% & 41.12 & 0.00\% & 31.64 & 0.00\% & 71.01 & 0.00\% \\ \hline
        24 & 20 & 44.15 & 0.37\% & 21.30 & 0.69\% & 41.31 & 0.46\% & 31.73 & 0.31\% & 71.20 & 0.26\% \\ \hline
        24 & 16 & 44.10 & 0.27\% & 21.32 & 0.81\% & 41.29 & 0.39\% & 31.83 & 0.60\% & 70.19 & -1.16\% \\ \hline
        24 & 12 & 43.74 & -0.57\% & 21.08 & -0.34\% & 40.97 & -0.38\% & 31.60 & -0.13\% & 69.99 & -1.44\% \\ \hline
        24 & 8 & 43.35 & -1.45\% & 20.67 & -2.27\% & 40.58 & -1.32\% & 31.29 & -1.11\% & 72.88 & 2.63\% \\ \hline
        24 & 4 & 41.42 & -5.84\% & 19.49 & -7.88\% & 38.78 & -5.69\% & 30.35 & -4.06\% & 70.39 & -0.89\% \\ \hline
        20 & 24 & 44.10 & 0.26\% & 21.13 & -0.12\% & 41.28 & 0.38\% & 31.58 & -0.17\% & 71.04 & 0.04\% \\ \hline
        16 & 24 & 43.76 & -0.52\% & 20.83 & -1.53\% & 40.92 & -0.49\% & 31.22 & -1.31\% & 71.59 & 0.80\% \\ \hline
        12 & 24 & 43.33 & -1.50\% & 20.53 & -2.94\% & 40.43 & -1.68\% & 30.82 & -2.58\% & 73.28 & 3.20\% \\ \hline
        8 & 24 & 42.46 & -3.48\% & 19.74 & -6.67\% & 39.64 & -3.60\% & 29.98 & -5.23\% & 73.47 & 3.46\% \\ \hline
        4 & 24 & 41.25 & -6.23\% & 18.68 & -11.69\% & 38.30 & -6.86\% & 28.78 & -9.04\% & 76.05 & 7.08\% \\ \hline
        20 & 20 & 44.10 & 0.25\% & 21.23 & 0.34\% & 41.25 & 0.32\% & 31.65 & 0.05\% & 70.90 & -0.16\% \\ \hline
        16 & 16 & 43.69 & -0.67\% & 20.90 & -1.19\% & 40.86 & -0.64\% & 31.30 & -1.06\% & 71.85 & 1.18\% \\ \hline
        12 & 12 & 42.81 & -2.67\% & 20.13 & -4.84\% & 39.97 & -2.80\% & 30.58 & -3.33\% & 72.81 & 2.53\% \\ \hline
        8 & 8 & 40.57 & -7.78\% & 18.47 & -12.70\% & 37.82 & -8.04\% & 28.96 & -8.46\% & 73.39 & 3.34\% \\ \hline
        4 & 4 & 36.11 & -17.91\% & 15.51 & -26.68\% & 33.48 & -18.59\% & 26.30 & -16.88\% & 68.58 & -3.43\% \\ \hline
    \end{tabular}
    \label{tab:asym-large-cnndm}
\end{table*}
\begin{table*}[!ht]
    \centering
    \caption{The relation between pruning asymmetry and symmetry for a FLAN-T5 small model on the Query Independent Web Snippets Abstractive Summarization Dataset}
    \small
    \begin{tabular}{|l|l|l|l|l|l|l|l|l|l|l|l|}
    \hline
        $l_{enc}$ & $l_{dec}$ & R-1 & Impact & R-2 & Impact & RSL & Impact & R-L & Impact & GenL & Impact \\ \hline
        8 & 8 & 50.22 & 100.00\% & 29.03 & 100.00\% & 45.87 & 100.00\% & 40.19 & 100.00\% & 62.79 & 100.00\% \\ \hline
        8 & 6 & 50.20 & 99.96\% & 28.90 & 99.55\% & 45.80 & 99.83\% & 40.45 & 100.65\% & 62.81 & 100.03\% \\ \hline
        8 & 5 & 49.74 & 99.04\% & 28.56 & 98.40\% & 45.55 & 99.30\% & 40.27 & 100.20\% & 62.68 & 99.83\% \\ \hline
        8 & 4 & 48.59 & 96.74\% & 27.94 & 96.24\% & 44.65 & 97.33\% & 39.27 & 97.70\% & 62.67 & 99.82\% \\ \hline
        8 & 2 & 45.36 & 90.32\% & 24.85 & 85.61\% & 41.38 & 90.21\% & 36.92 & 91.87\% & 62.68 & 99.84\% \\ \hline
        8 & 1 & 34.47 & 68.64\% & 15.41 & 53.08\% & 31.00 & 67.58\% & 27.68 & 68.88\% & 61.68 & 98.24\% \\ \hline
        6 & 8 & 49.32 & 98.21\% & 27.92 & 96.17\% & 44.72 & 97.48\% & 39.10 & 97.28\% & 62.90 & 100.18\% \\ \hline
        5 & 8 & 49.08 & 97.72\% & 27.75 & 95.60\% & 44.29 & 96.56\% & 38.76 & 96.45\% & 62.87 & 100.13\% \\ \hline
        4 & 8 & 46.40 & 92.39\% & 25.20 & 86.82\% & 41.81 & 91.14\% & 36.71 & 91.34\% & 62.74 & 99.93\% \\ \hline
        2 & 8 & 45.08 & 89.77\% & 23.67 & 81.55\% & 40.44 & 35.31\% & 35.31 & 87.85\% & 62.82 & 100.06\% \\ \hline
        1 & 8 & 39.81 & 79.26\% & 18.23 & 62.79\% & 35.39 & 77.14\% & 29.97 & 74.56\% & 62.83 & 100.07\% \\ \hline
        6 & 6 & 48.47 & 96.51\% & 26.82 & 92.38\% & 43.88 & 95.66\% & 38.38 & 95.49\% & 62.81 & 100.04\% \\ \hline
        5 & 5 & 47.55 & 94.68\% & 26.62 & 91.72\% & 43.13 & 94.02\% & 37.99 & 94.51\% & 62.67 & 99.81\% \\ \hline
        4 & 4 & 42.33 & 84.28\% & 23.12 & 79.64\% & 39.89 & 86.95\% & 33.39 & 83.08\% & 62.71 & 99.88\% \\ \hline
        2 & 2 & 39.69 & 79.02\% & 19.14 & 65.92\% & 35.49 & 77.36\% & 30.90 & 76.89\% & 62.79 & 100.00\% \\ \hline
        1 & 1 & 22.98 & 45.75\% & 6.09 & 20.99\% & 20.52 & 44.74\% & 18.36 & 45.69\% & 61.90 & 98.58\% \\ \hline
    \end{tabular}
    \label{tab:asym-small-qiws}
\end{table*}

\begin{table*}[!ht]
    \centering
    \caption{The relation between pruning asymmetry and symmetry for a FLAN-T5 base model on the Query Independent Web Snippets Abstractive Summarization Dataset}
    \small
    \begin{tabular}{|l|l|l|l|l|l|l|l|l|l|l|l|}
    \hline
        $l_{enc}$ & $l_{dec}$ & R-1 & Impact & R-2 & Impact & RSL & Impact & R-L & Impact & GenL & Impact \\ \hline
        12 & 12 & 54.84 & 0.00\% & 34.19 & 0.00\% & 50.38 & 0.00\% & 44.68 & 0.00\% & 62.91 & 0.00\% \\ \hline
        12 & 10 & 55.02 & 0.33\% & 34.00 & -0.56\% & 50.20 & -0.35\% & 44.67 & -0.02\% & 62.79 & -0.19\% \\ \hline
        12 & 8 & 55.97 & 2.05\% & 34.50 & 0.91\% & 51.12 & 1.48\% & 44.90 & 0.48\% & 62.75 & -0.24\% \\ \hline
        12 & 6 & 54.54 & -0.55\% & 33.70 & -1.42\% & 49.94 & -0.87\% & 44.19 & -1.11\% & 62.81 & -0.16\% \\ \hline
        12 & 4 & 52.64 & -4.01\% & 31.93 & -6.62\% & 47.28 & -6.16\% & 42.98 & -3.81\% & 62.85 & -0.09\% \\ \hline
        12 & 2 & 49.02 & -10.61\% & 28.05 & -17.97\% & 44.98 & -10.71\% & 40.36 & -9.68\% & 62.89 & -0.02\% \\ \hline
        10 & 12 & 54.23 & -1.11\% & 33.57 & -1.82\% & 49.93 & -0.89\% & 44.00 & -1.52\% & 62.87 & -0.05\% \\ \hline
        8 & 12 & 54.02 & -1.50\% & 33.06 & -3.31\% & 49.49 & -1.76\% & 43.80 & -1.96\% & 62.85 & -0.09\% \\ \hline
        6 & 12 & 48.74 & -11.13\% & 32.23 & -5.72\% & 48.74 & -3.26\% & 42.92 & -3.95\% & 62.82 & -0.14\% \\ \hline
        4 & 12 & 47.93 & -12.61\% & 27.47 & -19.65\% & 46.21 & -8.28\% & 39.77 & -11.00\% & 62.79 & -0.19\% \\ \hline
        2 & 12 & 47.45 & -13.48\% & 25.57 & -25.22\% & 43.20 & -14.26\% & 37.69 & -15.66\% & 62.77 & -0.22\% \\ \hline
        10 & 10 & 54.25 & -1.08\% & 32.88 & -3.82\% & 49.51 & -1.72\% & 43.24 & -3.23\% & 62.82 & -0.13\% \\ \hline
        8 & 8 & 53.89 & -1.73\% & 32.81 & -4.04\% & 49.32 & -2.10\% & 43.77 & -2.04\% & 62.82 & -0.14\% \\ \hline
        6 & 6 & 50.26 & -8.34\% & 28.70 & -16.05\% & 45.62 & -9.45\% & 40.05 & -10.37\% & 62.82 & -0.13\% \\ \hline
        4 & 4 & 47.77 & -12.89\% & 26.53 & -22.40\% & 43.34 & -13.97\% & 37.85 & -15.29\% & 62.84 & -0.10\% \\ \hline
        2 & 2 & 39.59 & -27.80\% & 19.64 & -42.57\% & 35.80 & -28.95\% & 31.38 & -29.78\% & 62.85 & -0.09\% \\ \hline
    \end{tabular}
    \label{tab:asym-base-qiws}
\end{table*}
\begin{table*}[!ht]
    \centering
    \caption{The relation between pruning asymmetry and symmetry for a FLAN-T5 large model on the Query Independent Web Snippets Abstractive Summarization Dataset}
    \small\begin{tabular}{|l|l|l|l|l|l|l|l|l|l|l|l|}
    \hline
        $l_{enc}$ & $l_{dec}$ & R-1 & Impact & R-2 & Impact & RSL & Impact & R-L & Impact & GenL & Impact \\ \hline
        24 & 24 & 57.81 & 100.00\% & 37.37 & 100.00\% & 53.14 & 100.00\% & 48.16 & 100.00\% & 62.85 & 100.00\% \\ \hline
        24 & 20 & 58.21 & 100.69\% & 37.59 & 100.59\% & 53.44 & 100.58\% & 48.46 & 100.62\% & 62.80 & 99.91\% \\ \hline
        24 & 16 & 57.25 & 99.04\% & 36.56 & 97.84\% & 52.71 & 99.19\% & 47.71 & 99.06\% & 62.83 & 99.97\% \\ \hline
        24 & 12 & 56.78 & 98.21\% & 35.74 & 95.64\% & 52.34 & 98.49\% & 46.81 & 97.18\% & 62.78 & 99.88\% \\ \hline
        24 & 8 & 56.19 & 97.19\% & 35.13 & 94.01\% & 51.59 & 97.08\% & 45.68 & 94.85\% & 62.79 & 99.90\% \\ \hline
        24 & 4 & 54.53 & 94.32\% & 33.69 & 90.15\% & 50.00 & 94.10\% & 44.65 & 92.71\% & 62.83 & 99.97\% \\ \hline
        20 & 24 & 57.34 & 99.19\% & 36.39 & 97.38\% & 52.66 & 99.10\% & 47.28 & 98.18\% & 62.81 & 99.93\% \\ \hline
        16 & 24 & 56.26 & 97.33\% & 35.90 & 96.07\% & 51.04 & 96.04\% & 46.82 & 97.22\% & 62.81 & 99.93\% \\ \hline
        12 & 24 & 55.31 & 95.67\% & 34.22 & 91.58\% & 50.60 & 95.23\% & 45.11 & 93.66\% & 62.88 & 100.04\% \\ \hline
        8 & 24 & 54.80 & 94.79\% & 33.42 & 89.43\% & 49.95 & 94.00\% & 44.11 & 91.59\% & 62.70 & 99.76\% \\ \hline
        4 & 24 & 51.40 & 88.92\% & 30.31 & 81.11\% & 46.49 & 87.48\% & 41.12 & 85.38\% & 62.70 & 99.75\% \\ \hline
        20 & 20 & 56.81 & 98.28\% & 36.32 & 97.20\% & 52.21 & 98.25\% & 46.82 & 97.21\% & 62.69 & 99.74\% \\ \hline
        16 & 16 & 56.10 & 97.05\% & 35.98 & 96.29\% & 51.05 & 96.07\% & 45.89 & 95.28\% & 62.71 & 99.76\% \\ \hline
        12 & 12 & 54.16 & 93.70\% & 33.00 & 88.31\% & 49.58 & 93.31\% & 44.80 & 93.02\% & 62.77 & 99.87\% \\ \hline
        8 & 8 & 51.77 & 89.55\% & 30.78 & 82.38\% & 47.31 & 89.03\% & 41.32 & 85.79\% & 62.73 & 99.81\% \\ \hline
        4 & 4 & 45.70 & 79.06\% & 22.77 & 60.94\% & 41.36 & 77.84\% & 36.09 & 74.94\% & 62.70 & 99.76\% \\ \hline
    \end{tabular}
    \label{tab:asym-large-qiws}
\end{table*}
\begin{table*}[!ht]
    \centering
    \caption{The relation between pruning asymmetry and symmetry for a FLAN-T5 small model on the Extreme Summarization (XSUM) Abstractive Summarization Dataset}
    \small
    \begin{tabular}{|l|l|l|l|l|l|l|l|l|l|l|l|}
    \hline
        $l_{enc}$ & $l_{dec}$ & R-1 & Impact & R-2 & Impact & RSL & Impact & R-L & Impact & GenL & Impact \\ \hline
        8 & 8 & 33.27 & 0.00\% & 11.09 & 0.00\% & 26.17 & 0.00\% & 26.17 & 0.00\% & 28.01 & 0.00\% \\ \hline
        8 & 6 & 33.79 & 1.56\% & 11.61 & 4.74\% & 26.73 & 2.14\% & 26.74 & 2.18\% & 27.79 & -0.78\% \\ \hline
        8 & 5 & 33.47 & 0.61\% & 11.43 & 3.12\% & 26.64 & 1.81\% & 26.65 & 1.83\% & 27.40 & -2.18\% \\ \hline
        8 & 3 & 33.04 & -0.69\% & 11.24 & 1.36\% & 26.26 & 0.36\% & 26.27 & 0.38\% & 28.08 & 0.26\% \\ \hline
        8 & 2 & 31.48 & -5.36\% & 10.53 & -5.02\% & 25.39 & -2.99\% & 25.38 & -3.01\% & 26.58 & -5.13\% \\ \hline
        8 & 1 & 23.16 & -30.39\% & 6.03 & -45.58\% & 19.02 & -27.32\% & 19.02 & -27.33\% & 36.68 & 30.93\% \\ \hline
        5 & 8 & 33.31 & 0.13\% & 11.18 & 0.82\% & 26.16 & -0.04\% & 26.16 & -0.06\% & 28.31 & 1.08\% \\ \hline
        5 & 8 & 32.55 & -2.15\% & 10.61 & -4.32\% & 25.50 & -2.55\% & 25.50 & -2.55\% & 28.35 & 1.19\% \\ \hline
        3 & 8 & 31.82 & -4.36\% & 10.11 & -8.84\% & 24.92 & -4.78\% & 24.92 & -4.77\% & 28.43 & 1.50\% \\ \hline
        2 & 8 & 29.65 & -10.87\% & 8.59 & -22.48\% & 23.02 & -12.02\% & 23.02 & -12.03\% & 27.90 & -0.39\% \\ \hline
        1 & 8 & 28.46 & -14.46\% & 7.70 & -30.57\% & 22.09 & -15.60\% & 22.09 & -15.59\% & 27.87 & -0.50\% \\ \hline
        6 & 6 & 32.50 & -2.29\% & 10.73 & -3.24\% & 25.67 & -1.90\% & 25.68 & -1.88\% & 28.07 & 0.19\% \\ \hline
        5 & 5 & 31.77 & -4.50\% & 10.19 & -8.04\% & 25.14 & -3.94\% & 25.14 & -3.95\% & 28.09 & 0.29\% \\ \hline
        3 & 3 & 30.42 & -8.57\% & 9.50 & -14.31\% & 24.16 & -7.66\% & 24.16 & -7.67\% & 27.91 & -0.38\% \\ \hline
        2 & 2 & 26.71 & -19.70\% & 7.31 & -34.09\% & 21.38 & -18.30\% & 21.38 & -18.31\% & 26.35 & -5.93\% \\ \hline
        1 & 1 & 19.54 & -41.26\% & 4.00 & -63.91\% & 16.00 & -38.86\% & 16.00 & -38.87\% & 35.73 & 27.54\% \\ \hline
    \end{tabular}
    \label{tab:asym-small-xsum}
\end{table*}

\begin{table*}[!ht]
    \centering
    \caption{The relation between pruning asymmetry and symmetry for a FLAN-T5 base model on the Extreme Summarization (XSUM) Abstractive Summarization Dataset}
    \small
    \begin{tabular}{|l|l|l|l|l|l|l|l|l|l|l|l|}
    \hline
        $l_{enc}$ & $l_{dec}$ & R-1 & Impact & R-2 & Impact & RSL & Impact & R-L & Impact & GenL & Impact \\ \hline
        12 & 12 & 38.78 & 0.00\% & 15.69 & 0.00\% & 31.14 & 0.00\% & 31.15 & 0.00\% & 25.92 & 0.00\% \\ \hline
        12 & 10 & 38.46 & -0.83\% & 15.27 & -2.65\% & 30.70 & -1.43\% & 30.71 & -1.42\% & 26.72 & 3.11\% \\ \hline
        12 & 8 & 38.11 & -1.72\% & 14.91 & -4.97\% & 30.34 & -2.59\% & 30.34 & -2.60\% & 27.64 & 6.65\% \\ \hline
        12 & 6 & 38.55 & -0.58\% & 15.40 & -1.83\% & 30.87 & -0.87\% & 30.88 & -0.87\% & 27.42 & 5.80\% \\ \hline
        12 & 4 & 38.04 & -1.91\% & 15.19 & -3.18\% & 30.63 & -1.64\% & 29.65 & -4.82\% & 26.40 & 1.85\% \\ \hline
        12 & 2 & 35.39 & -8.74\% & 13.73 & -12.47\% & 28.96 & -7.02\% & 28.96 & -7.03\% & 27.55 & 6.32\% \\ \hline
        10 & 12 & 39.04 & 0.68\% & 15.92 & 1.47\% & 31.22 & 0.24\% & 31.23 & 0.25\% & 26.89 & 3.75\% \\ \hline
        8 & 12 & 37.05 & -4.45\% & 14.10 & -10.09\% & 29.29 & -5.95\% & 29.30 & -5.93\% & 27.68 & 6.82\% \\ \hline
        6 & 12 & 36.45 & -6.01\% & 13.84 & -11.79\% & 28.96 & -7.02\% & 28.96 & -7.02\% & 27.21 & 4.99\% \\ \hline
        4 & 12 & 34.32 & -11.48\% & 12.10 & -22.88\% & 26.99 & -13.35\% & 26.99 & -13.34\% & 27.20 & 4.94\% \\ \hline
        2 & 12 & 31.88 & -17.78\% & 10.27 & -34.53\% & 24.85 & -20.21\% & 24.85 & -20.22\% & 28.22 & 8.88\% \\ \hline
        10 & 10 & 38.80 & 0.05\% & 15.72 & 0.22\% & 31.07 & -0.25\% & 31.08 & -0.23\% & 26.92 & 3.88\% \\ \hline
        8 & 8 & 37.21 & -4.04\% & 14.30 & -8.85\% & 29.55 & -5.13\% & 29.54 & -5.15\% & 27.40 & 5.72\% \\ \hline
        6 & 6 & 34.92 & -9.95\% & 12.44 & -20.68\% & 27.56 & -11.51\% & 27.57 & -11.50\% & 27.72 & 6.96\% \\ \hline
        4 & 4 & 32.48 & -16.24\% & 10.67 & -31.95\% & 25.49 & -18.15\% & 25.50 & -18.14\% & 27.98 & 7.98\% \\ \hline
        2 & 2 & 27.44 & -29.23\% & 7.74 & -50.65\% & 21.95 & -29.51\% & 21.96 & -29.52\% & 29.38 & 13.38\% \\ \hline
    \end{tabular}
    \label{tab:asym-base-xsum}
\end{table*}

\begin{table*}[!ht]
    \centering
    \caption{The relation between pruning asymmetry and symmetry for a FLAN-T5 large model on the Extreme Summarization (XSUM) Abstractive Summarization Dataset}
    \small
    \begin{tabular}{|l|l|l|l|l|l|l|l|l|l|l|l|}
    \hline
        $l_{enc}$ & $l_{dec}$ & R-1 & Impact & R-2 & Impact & RSL & Impact & R-L & Impact & GenL & Impact \\ \hline
        24 & 24 & 39.71 & 0.00\% & 16.34 & 0.00\% & 31.72 & 0.00\% & 31.72 & 0.01\% & 26.74 & 0.00\% \\ \hline
        24 & 20 & 43.18 & 8.74\% & 19.80 & 21.17\% & 35.21 & 11.01\% & 35.22 & 11.04\% & 25.91 & -3.10\% \\ \hline
        24 & 16 & 42.73 & 7.59\% & 19.30 & 18.10\% & 34.76 & 9.58\% & 34.76 & 9.59\% & 26.40 & -1.29\% \\ \hline
        24 & 12 & 42.34 & 6.61\% & 18.92 & 15.78\% & 34.52 & 8.84\% & 34.53 & 8.87\% & 25.49 & -4.68\% \\ \hline
        24 & 8 & 41.30 & 4.00\% & 17.96 & 9.94\% & 33.73 & 6.34\% & 33.75 & 6.39\% & 25.02 & -6.45\% \\ \hline
        24 & 4 & 39.55 & -0.40\% & 16.47 & 0.77\% & 32.25 & 1.66\% & 32.25 & 1.68\% & 26.30 & -1.64\% \\ \hline
        20 & 24 & 42.77 & 7.71\% & 19.43 & 18.90\% & 34.83 & 9.82\% & 34.84 & 9.83\% & 26.18 & -2.09\% \\ \hline
        16 & 24 & 41.55 & 4.63\% & 18.33 & 12.17\% & 33.64 & 6.05\% & 33.65 & 6.07\% & 26.33 & -1.53\% \\ \hline
        12 & 24 & 39.95 & 0.61\% & 16.90 & 3.40\% & 32.13 & 1.29\% & 32.14 & 1.31\% & 27.14 & -100.00\% \\ \hline
        8 & 24 & 37.57 & -5.39\% & 14.97 & -8.38\% & 29.94 & -5.61\% & 29.94 & -5.60\% & 25.99 & -100.00\% \\ \hline
        4 & 24 & 34.81 & -12.35\% & 12.52 & -23.36\% & 27.32 & -13.86\% & 27.32 & -13.86\% & 27.61 & -100.00\% \\ \hline
        20 & 20 & 42.48 & 6.98\% & 19.18 & 17.39\% & 34.62 & 9.13\% & 34.62 & 9.13\% & 25.84 & -3.36\% \\ \hline
        16 & 16 & 40.78 & 2.69\% & 17.56 & 7.44\% & 32.99 & 4.00\% & 33.00 & 4.02\% & 26.47 & -1.00\% \\ \hline
        12 & 12 & 38.94 & 6.98\% & 15.89 & -2.78\% & 31.21 & -1.61\% & 31.22 & -1.58\% & 26.59 & -0.57\% \\ \hline
        8 & 8 & 34.65 & -12.75\% & 12.15 & -25.65\% & 27.36 & -13.76\% & 27.36 & -13.73\% & 28.16 & 5.30\% \\ \hline
        4 & 4 & 29.82 & -24.91\% & 8.96 & -45.14\% & 23.59 & -25.62\% & 23.60 & -25.60\% & 28.10 & 5.09\% \\ \hline
    \end{tabular}
    \label{tab:asym-large-xsum}
\end{table*}
\subsection{Inference Benchmarks}
\label{sec:inference-benchmarks}
Detailed variations in latency measurements across batch size, scale, structural pruning, and asymmetry on performance on the QIWS, CNN/DM, and XSUM datasets can be found in tables \ref{tab:qiws-asym-inference-small},\ref{tab:qiws-asym-inference-base}, \ref{tab:qiws-asym-inference-large}, \ref{tab:cnndm-asym-inference-small},\ref{tab:cnndm-asym-inference-base}, \ref{tab:cnndm-asym-inference-large}, \ref{tab:xsum-asym-inference-large}, \ref{tab:xsum-asym-inference-small}, and  \ref{tab:xswum-asym-inference-base}.
\begin{table*}[!ht]
    \centering
    \caption{Role of model symmetry in inference efficiency on FLAN-T5 small model on the QIWS dataset}
    \small
    \begin{tabular}{|l|l|l|l|l|l|l|l|l|l|l|l|l|}
    \hline
         $l_{enc}$ & $l_{dec}$ & R-2 & Impact & BS 1 & STD & Speedup & BS 8 & STD  & Speedup & BS 16 & STD & Speedup \\ \hline
         8 & 8 & 29.03 & 0.00\% & 524 & 3.95 & 1.00 & 653 & 2.49 & 1.00 & 729 & 5.12 & 1.00 \\ \hline
        8 & 6 & 28.90 & -0.45\% & 406 & 1.28 & 1.29 & 514 & 5.02 & 1.27 & 583 & 2.47 & 1.25 \\ \hline
        8 & 5 & 28.56 & -1.60\% & 348 & 2.34 & 1.51 & 455 & 1.6 & 1.44 & 527 & 1.85 & 1.38 \\ \hline
        8 & 4 & 27.94 & -3.76\% & 293 & 3.35 & 1.79 & 394 & 6.32 & 1.66 & 469 & 2.65 & 1.55 \\ \hline
        8 & 2 & 24.85 & -14.39\% & 195 & 1.61 & 2.69 & 353 & 3.38 & 1.85 & 426 & 6.38 & 1.71 \\ \hline
        8 & 1 & 15.41 & -46.92\% & 132 & 0.959 & 3.97 & 211 & 2.82 & 3.09 & 389 & 2.94 & 1.87 \\ \hline
        6 & 8 & 27.92 & -3.83\% & 512 & 5.15 & 1.02 & 626 & 4.19 & 1.04 & 684 & 2.81 & 1.07 \\ \hline
        5 & 8 & 27.75 & -4.40\% & 508 & 3.56 & 1.03 & 617 & 4.91 & 1.06 & 666 & 4.16 & 1.09 \\ \hline
        4 & 8 & 25.20 & -13.18\% & 514 & 3.55 & 1.02 & 603 & 4.52 & 1.08 & 639 & 2.08 & 1.14 \\ \hline
        2 & 8 & 23.67 & -18.45\% & 514 & 514 & 1.02 & 585 & 5.36 & 1.12 & 608 & 4.45 & 1.20 \\ \hline
        1 & 8 & 18.23 & -37.21\% & 510 & 5.81 & 1.03 & 574 & 4.21 & 1.14 & 595 & 7.06 & 1.23 \\ \hline
        6 & 6 & 26.82 & -7.62\% & 407 & 5.26 & 1.29 & 496 & 8.77 & 1.32 & 548 & 1.97 & 1.33 \\ \hline
        5 & 5 & 26.62 & -8.28\% & 346 & 6.84 & 1.51 & 430 & 3.54 & 1.52 & 480 & 12.4 & 1.52 \\ \hline
        4 & 4 & 23.12 & -20.36\% & 375 & 4.25 & 1.40 & 441 & 6.92 & 1.48 & 478 & 10.6 & 1.53 \\ \hline
        2 & 2 & 19.14 & -34.08\% & 402 & 2.05 & 1.30 & 452 & 9.84 & 1.44 & 476 & 8.29 & 1.53 \\ \hline
        1 & 1 & 6.09 & -79.01\% & 134 & 6.2 & 3.91 & 527 & 3.03 & 1.24 & 549 & 13.4 & 1.33 \\ \hline
    \end{tabular}
    \label{tab:qiws-asym-inference-small}
\end{table*}
\begin{table*}[!ht]
    \centering
    \caption{Role of model symmetry in inference efficiency on FLAN-T5 base model on the QIWS dataset}
    \small
    \begin{tabular}{|l|l|l|l|l|l|l|l|l|l|l|l|l|}
    \hline
         $l_{enc}$ & $l_{dec}$ & R-2 & Impact & BS 1 & STD & Speedup & BS 8 & STD  & Speedup & BS 16 & STD & Speedup \\ \hline
         12 & 12 & 34.19 & 0.00\% & 746 & 11 & 1.00 & 1060 & 2.84 & 1.00 & 1310 & 6.8 & 1.00 \\ \hline
        12 & 10 & 34.00 & -0.56\% & 625 & 3.27 & 1.19 & 943 & 4.69 & 1.12 & 1200 & 4.8 & 1.09 \\ \hline
        12 & 8 & 34.50 & 0.91\% & 523 & 2.19 & 1.43 & 814 & 4.23 & 1.30 & 1070 & 5.34 & 1.22 \\ \hline
        12 & 6 & 33.70 & -1.42\% & 425 & 1.92 & 1.76 & 652 & 3.39 & 1.63 & 970 & 4.79 & 1.35 \\ \hline
        12 & 4 & 31.93 & -6.62\% & 350 & 1.32 & 2.13 & 510 & 3.1 & 2.08 & 815 & 2 & 1.61 \\ \hline
        12 & 2 & 28.05 & -17.97\% & 202 & 1.41 & 3.69 & 451 & 2.92 & 2.35 & 762 & 0.911 & 1.72 \\ \hline
        10 & 12 & 33.57 & -1.82\% & 710 & 6.2 & 1.05 & 995 & 2.74 & 1.07 & 1290 & 4.2 & 1.02 \\ \hline
        8 & 12 & 33.06 & -3.31\% & 690 & 5.72 & 1.08 & 953 & 5.72 & 1.11 & 1270 & 4.3 & 1.03 \\ \hline
        6 & 12 & 32.23 & -5.72\% & 716 & 8 & 1.04 & 944 & 7.22 & 1.12 & 1080 & 5.29 & 1.21 \\ \hline
        4 & 12 & 27.47 & -19.65\% & 710 & 1.75 & 1.05 & 911 & 10.1 & 1.16 & 1,000 & 8.84 & 1.31 \\ \hline
        2 & 12 & 25.57 & -25.22\% & 706 & 5.4 & 1.06 & 862 & 7.11 & 1.23 & 921 & 7.04 & 1.42 \\ \hline
        10 & 10 & 32.88 & -3.82\% & 633 & 11.6 & 1.18 & 915 & 11 & 1.16 & 1120 & 5.51 & 1.17 \\ \hline
        8 & 8 & 32.81 & -4.04\% & 512 & 4.98 & 1.46 & 737 & 9.78 & 1.44 & 911 & 4.98 & 1.44 \\ \hline
        6 & 6 & 28.70 & -16.05\% & 401 & 3.16 & 1.86 & 572 & 4.73 & 1.85 & 702 & 1.57 & 1.87 \\ \hline
        4 & 4 & 26.53 & -22.40\% & 301 & 2.92 & 2.48 & 415 & 3.01 & 2.55 & 509 & 0.997 & 2.57 \\ \hline
        2 & 2 & 19.64 & -42.57\% & 189 & 1.98 & 3.95 & 312 & 2.88 & 3.40 & 389 & 0.892 & 3.37 \\ \hline
    \end{tabular}
    \label{tab:qiws-asym-inference-base}
\end{table*}
\begin{table*}[!ht]
    \centering
    \caption{Role of model symmetry in inference efficiency on FLAN-T5 large model on the QIWS dataset}
    \small
    \small
    \begin{tabular}{|l|l|l|l|l|l|l|l|l|l|l|l|l|}
    \hline
         $l_{enc}$ & $l_{dec}$ & R-2 & Impact & BS 1 & STD & Speedup & BS 8 & STD  & Speedup & BS 16 & STD & Speedup \\ \hline
         24 & 24 & 37.37 & 0.00\% & 1430 & 6.08 & 1.00 & 2240 & 4.81 & 1.00 & 3320 & 1.02 & 1.00 \\ \hline
        24 & 20 & 37.59 & 0.59\% & 1210 & 4.73 & 1.18 & 1990 & 6.89 & 1.13 & 3010 & 2.63 & 1.10 \\ \hline
        24 & 16 & 36.56 & -2.16\% & 1000 & 2.70 & 1.43 & 1750 & 5.92 & 1.28 & 2710 & 1.57 & 1.23 \\ \hline
        24 & 12 & 35.74 & -4.36\% & 795 & 6.61 & 1.80 & 1510 & 10.40 & 1.48 & 2400 & 1.59 & 1.38 \\ \hline
        24 & 8 & 35.13 & -5.99\% & 585 & 4.99 & 2.44 & 1260 & 7.14 & 1.78 & 2090 & 7.17 & 1.59 \\ \hline
        24 & 4 & 33.69 & -9.85\% & 373 & 1.16 & 3.83 & 1030 & 10.50 & 2.17 & 1790 & 1.72 & 1.85 \\ \hline
        20 & 24 & 36.39 & -2.62\% & 1410 & 3.66 & 1.01 & 2130 & 10.90 & 1.05 & 3090 & 5.98 & 1.07 \\ \hline
        16 & 24 & 35.90 & -3.93\% & 1395 & 3.52 & 1.03 & 2060 & 9.89 & 1.09 & 2880 & 3.32 & 1.15 \\ \hline
        12 & 24 & 34.22 & -8.42\% & 1380 & 5.20 & 1.04 & 1900 & 9.65 & 1.18 & 2630 & 0.81 & 1.26 \\ \hline
        8 & 24 & 33.42 & -10.57\% & 1370 & 5.49 & 1.04 & 1790 & 19.00 & 1.25 & 2400 & 1.34 & 1.38 \\ \hline
        4 & 24 & 30.31 & -18.89\% & 1350 & 7.33 & 1.06 & 1670 & 5.30 & 1.34 & 2170 & 2.79 & 1.53 \\ \hline
        20 & 20 & 36.32 & -2.80\% & 1200 & 5.37 & 1.19 & 1880 & 7.89 & 1.19 & 2780 & 1.15 & 1.19 \\ \hline
        16 & 16 & 35.98 & -3.71\% & 1020 & 3.49 & 1.40 & 1530 & 5.62 & 1.46 & 2230 & 1.80 & 1.49 \\ \hline
        12 & 12 & 33.00 & -11.69\% & 749 & 5.30 & 1.91 & 1160 & 2.94 & 1.93 & 1710 & 0.89 & 1.94 \\ \hline
        8 & 8 & 30.78 & -17.62\% & 650 & 3.32 & 2.20 & 970 & 2.78 & 2.31 & 1550 & 0.79 & 2.14 \\ \hline
        4 & 4 & 22.77 & -39.06\% & 585 & 2.23 & 2.44 & 890 & 3.21 & 2.52 & 1450 & 0.92 & 2.29 \\ \hline
    \end{tabular}
    \label{tab:qiws-asym-inference-large}
\end{table*}

\begin{table*}[!ht]
    \centering
    \caption{Role of model symmetry in inference efficiency on FLAN-T5 small model on the CNNDM dataset}
    \small
    \begin{tabular}{|l|l|l|l|l|l|l|l|l|l|l|l|l|}
    \hline
         $l_{enc}$ & $l_{dec}$ & R-2 & Impact & BS 1 & STD & Speedup & BS 8 & STD  & Speedup & BS 16 & STD & Speedup\\ \hline
         8 & 8 & 17.55 & 0.00\% & 138 & 5.05 & 1.00 & 230 & 7.61 & 1.00 & 330 & 3.71 & 1.00 \\ \hline
        8 & 6 & 17.68 & 0.74\% & 133 & 0.292 & 1.04 & 211 & 0.425 & 1.09 & 300 & 0.954 & 1.10 \\ \hline
        8 & 5 & 17.27 & -1.64\% & 116 & 0.196 & 1.19 & 193 & 0.448 & 1.19 & 279 & 0.537 & 1.18 \\ \hline
        8 & 4 & 16.40 & -6.57\% & 98.1 & 0.242 & 1.41 & 174 & 0.153 & 1.32 & 259 & 0.424 & 1.27 \\ \hline
        8 & 2 & 15.35 & -12.58\% & 63.2 & 0.207 & 2.18 & 137 & 0.1 & 1.68 & 218 & 0.303 & 1.51 \\ \hline
        8 & 1 & 11.33 & -35.43\% & 45.7 & 0.106 & 3.02 & 118 & 0.0827 & 1.95 & 198 & 0.148 & 1.67 \\ \hline
        6 & 8 & 17.69 & 0.81\% & 166 & 0.303 & 0.83 & 230 & 1.42 & 1.00 & 303 & 1.06 & 1.09 \\ \hline
        5 & 8 & 17.35 & -1.16\% & 165 & 0.267 & 0.84 & 219 & 0.521 & 1.05 & 283 & 1.13 & 1.17 \\ \hline
        4 & 8 & 16.80 & -4.30\% & 164 & 0.185 & 0.84 & 211 & 0.89 & 1.09 & 265 & 1.85 & 1.25 \\ \hline
        2 & 8 & 15.54 & -11.49\% & 162 & 332 & 0.85 & 191 & 0.332 & 1.20 & 226 & 625 & 1.46 \\ \hline
        1 & 8 & 13.31 & -24.17\% & 161 & 0.626 & 0.86 & 180 & 0.423 & 1.28 & 206 & 0.55 & 1.60 \\ \hline
        6 & 6 & 17.07 & -2.77\% & 131 & 0.617 & 1.05 & 192 & 0.247 & 1.20 & 261 & 0.768 & 1.26 \\ \hline
        5 & 5 & 16.20 & -7.72\% & 113 & 0.306 & 1.22 & 164 & 0.642 & 1.40 & 220 & 1.36 & 1.50 \\ \hline
        4 & 4 & 14.91 & -15.05\% & 95.1 & 0.0955 & 1.45 & 135 & 0.21 & 1.70 & 182 & 0.268 & 1.81 \\ \hline
        2 & 2 & 11.97 & -31.83\% & 57.8 & 0.27 & 2.39 & 78.9 & 0.078 & 2.92 & 103 & 0.238 & 3.20 \\ \hline
        1 & 1 & 6.05 & -65.55\% & 39.1 & 0.136 & 3.53 & 50.2 & 0.132 & 4.58 & 63.4 & 0.0845 & 5.21 \\ \hline
    \end{tabular}
    \label{tab:cnndm-asym-inference-small}
\end{table*}

\begin{table*}[!ht]
    \centering
    \caption{Role of model symmetry in inference efficiency on FLAN-T5 base model on the CNNDM dataset}
    \small
    \begin{tabular}{|l|l|l|l|l|l|l|l|l|l|l|l|l|l|}
    \hline
         $l_{enc}$ & $l_{dec}$ & R-2 & Impact & BS 1 & STD & Speedup & BS 8 & STD  & Speedup & BS 16 & STD & Speedup \\ \hline
         12 & 12 & 19.77 & 0.00\% & 199 & 3.74 & 1.00 & 550 & 3.81 & 1.00 & 931 & 2.09 & 1.00 \\ \hline
        12 & 10 & 19.92 & 0.76\% & 179 & 3.31 & 1.11 & 524 & 16.2 & 1.05 & 889 & 4.41 & 1.05 \\ \hline
        12 & 8 & 19.85 & 0.42\% & 155 & 4.50 & 1.28 & 493 & 14 & 1.12 & 884 & 3.61 & 1.05 \\ \hline
        12 & 6 & 18.85 & -4.63\% & 126 & 1.95 & 1.58 & 449 & 5.88 & 1.22 & 800 & 4.59 & 1.16 \\ \hline
        12 & 4 & 18.68 & -5.49\% & 99.2 & 1.02 & 2.01 & 405 & 1.41 & 1.36 & 737 & 5.06 & 1.26 \\ \hline
        12 & 2 & 16.48 & -16.62\% & 75.3 & 0.85 & 2.64 & 372 & 1.98 & 1.48 & 697 & 4.55 & 1.34 \\ \hline
        10 & 12 & 19.92 & 0.76\% & 198 & 4.75 & 1.01 & 495 & 14.5 & 1.11 & 811 & 1.18 & 1.15 \\ \hline
        8 & 12 & 19.67 & -0.50\% & 196 & 3.72 & 1.02 & 441 & 7.82 & 1.25 & 715 & 4.39 & 1.30 \\ \hline
        6 & 12 & 18.85 & -4.63\% & 187 & 4.81 & 1.06 & 396 & 13.3 & 1.39 & 613 & 9.45 & 1.52 \\ \hline
        4 & 12 & 18.22 & -7.86\% & 183 & 3.54 & 1.09 & 330 & 5.04 & 1.67 & 509 & 2.1 & 1.83 \\ \hline
        2 & 12 & 17.06 & -13.73\% & 176 & 3.52 & 1.13 & 272 & 1.79 & 2.02 & 400 & 3.25 & 2.33 \\ \hline
        10 & 10 & 19.72 & -0.26\% & 171 & 3.21 & 1.16 & 462 & 11.9 & 1.19 & 776 & 4.62 & 1.20 \\ \hline
        8 & 8 & 19.17 & -3.01\% & 141 & 2.97 & 1.41 & 37 & 12.1 & 14.86 & 628 & 6.48 & 1.48 \\ \hline
        6 & 6 & 17.46 & -11.71\% & 109 & 1.71 & 1.83 & 281 & 2.61 & 1.96 & 478 & 3.55 & 1.95 \\ \hline
        4 & 4 & 15.87 & -19.74\% & 82.5 & 1.24 & 2.41 & 198 & 1.71 & 2.78 & 329 & 0.74 & 2.83 \\ \hline
        2 & 2 & 12.23 & -38.12\% & 50.7 & 1.30 & 3.93 & 112 & 2.59 & 4.91 & 178 & 0.557 & 5.23 \\ \hline
    \end{tabular}
    \label{tab:cnndm-asym-inference-base}
\end{table*}

\begin{table*}[!ht]
    \centering
    \caption{Role of model symmetry in inference efficiency on FLAN-T5 LARGE model on the CNNDM dataset}
    \small
    \begin{tabular}{|l|l|l|l|l|l|l|l|l|l|l|l|l|}
    \hline
         $l_{enc}$ & $l_{dec}$ & R-2 & Impact & BS 1 & STD & Speedup & BS 8 & STD  & Speedup & BS 16 & STD & Speedup \\ \hline
         24 & 24 & 21.15 & 0.00\% & 445 & 2.35 & 1.00 & 1480 & 20.1 & 1.00 & 2700 & 7.22 & 1.00 \\ \hline
        24 & 20 & 21.30 & 0.69\% & 390 & 33.7 & 1.14 & 1390 & 4.24 & 1.06 & 2590 & 7.7 & 1.04 \\ \hline
        24 & 16 & 21.32 & 0.81\% & 335 & 13.9 & 1.33 & 1330 & 7.7 & 1.11 & 2470 & 7.42 & 1.09 \\ \hline
        24 & 12 & 21.08 & -0.34\% & 270 & 3.28 & 1.65 & 1250 & 11 & 1.18 & 2340 & 6.68 & 1.15 \\ \hline
        24 & 8 & 20.67 & -2.27\% & 219 & 8.67 & 2.03 & 1180 & 8.17 & 1.25 & 2220 & 4.25 & 1.22 \\ \hline
        24 & 4 & 19.49 & -7.88\% & 165 & 1.81 & 2.70 & 1090 & 6.6 & 1.36 & 2090 & 9.15 & 1.29 \\ \hline
        20 & 24 & 21.13 & -0.12\% & 418 & 13.8 & 1.06 & 1320 & 15.3 & 1.12 & 2400 & 7.26 & 1.13 \\ \hline
        16 & 24 & 20.83 & -1.53\% & 421 & 16.8 & 1.06 & 1150 & 16 & 1.29 & 2080 & 6.07 & 1.30 \\ \hline
        12 & 24 & 20.53 & -2.94\% & 391 & 12.5 & 1.14 & 1000 & 21.7 & 1.48 & 1750 & 8.18 & 1.54 \\ \hline
        8 & 24 & 19.74 & -6.67\% & 373 & 13.1 & 1.19 & 882 & 6.92 & 1.68 & 1430 & 4.79 & 1.89 \\ \hline
        4 & 24 & 18.68 & -11.69\% & 350 & 4.32 & 1.27 & 670 & 15 & 2.21 & 1110 & 3.21 & 2.43 \\ \hline
        20 & 20 & 21.23 & 0.34\% & 359 & 4.3 & 1.24 & 1240 & 15.3 & 1.19 & 2260 & 6.73 & 1.19 \\ \hline
        16 & 16 & 20.90 & -1.19\% & 1289 & 2.5 & 0.35 & 994 & 21.6 & 1.49 & 1820 & 4.27 & 1.48 \\ \hline
        12 & 12 & 20.13 & -4.84\% & 229 & 12.1 & 1.94 & 756 & 12.6 & 1.96 & 1370 & 4.6 & 1.97 \\ \hline
        8 & 8 & 18.47 & -12.70\% & 160 & 31.8 & 2.78 & 513 & 2.55 & 2.88 & 926 & 7.24 & 2.92 \\ \hline
        4 & 4 & 15.51 & -26.68\% & 89.7 & 0.588 & 4.96 & 267 & 2.14 & 5.54 & 479 & 4.3 & 5.64 \\ \hline
    \end{tabular}
    \label{tab:cnndm-asym-inference-large}
\end{table*}

\begin{table*}[!ht]
    \centering
    \caption{Role of model symmetry in inference efficiency on FLAN-T5 small model on the XSUM dataset}
    \small
    \begin{tabular}{|l|l|l|l|l|l|l|l|l|l|l|l|l|}
    \hline
         $l_{enc}$ & $l_{dec}$ & R-2 & Impact & BS 1 & STD & Speedup & BS 8 & STD  & Speedup & BS 16 & STD & Speedup\\ \hline
         8 & 8 & 11.09 & 0.00\% & 135 & 2.73 & 1.00 & 227 & 3.51 & 1.00 & 332 & 1.91 & 1.00 \\ \hline
        8 & 6 & 11.61 & 4.74\% & 108 & 1.70 & 1.25 & 196 & 1.94 & 1.16 & 303 & 7.95 & 1.10 \\ \hline
        8 & 5 & 11.43 & 3.12\% & 94.1 & 3.02 & 1.43 & 183 & 3.43 & 1.24 & 281 & 6.77 & 1.18 \\ \hline
        8 & 4 & 11.24 & 1.36\% & 82.7 & 2.66 & 1.63 & 168 & 2.33 & 1.35 & 263 & 2.24 & 1.26 \\ \hline
        8 & 2 & 10.53 & -5.02\% & 55.8 & 1.72 & 2.42 & 141 & 1.53 & 1.61 & 234 & 5.01 & 1.42 \\ \hline
        8 & 1 & 6.03 & -45.58\% & 41.1 & 0.64 & 3.28 & 124 & 0.414 & 1.83 & 215 & 4.69 & 1.54 \\ \hline
        6 & 8 & 11.18 & 0.82\% & 133 & 3.51 & 1.02 & 204 & 3.63 & 1.11 & 295 & 5.72 & 1.13 \\ \hline
        5 & 8 & 10.61 & -4.32\% & 134 & 3.42 & 1.01 & 193 & 3.76 & 1.18 & 273 & 10.4 & 1.22 \\ \hline
        4 & 8 & 10.11 & -8.84\% & 130 & 2.77 & 1.04 & 185 & 13.6 & 1.23 & 245 & 6.45 & 1.36 \\ \hline
        2 & 8 & 8.59 & -22.48\% & 126 & 4.77 & 1.07 & 163 & 6 & 1.39 & 203 & 4.1 & 1.64 \\ \hline
        1 & 8 & 7.70 & -30.57\% & 126 & 3.38 & 1.07 & 148 & 2.02 & 1.53 & 180 & 2.85 & 1.84 \\ \hline
        6 & 6 & 10.73 & -3.24\% & 104 & 0.45 & 1.30 & 178 & 3.24 & 1.28 & 254 & 2.37 & 1.31 \\ \hline
        5 & 5 & 10.19 & -8.04\% & 91.6 & 2.10 & 1.47 & 151 & 1.78 & 1.50 & 219 & 10.3 & 1.52 \\ \hline
        4 & 4 & 9.50 & -14.31\% & 79 & 3.38 & 1.71 & 124 & 2.42 & 1.83 & 178 & 1.59 & 1.87 \\ \hline
        2 & 2 & 7.31 & -34.09\% & 49.5 & 2.56 & 2.73 & 74.8 & 1.9 & 3.03 & 101 & 0.719 & 3.29 \\ \hline
        1 & 1 & 4.00 & -63.91\% & 32 & 1.25 & 4.22 & 48.7 & 2.11 & 4.66 & 61.9 & 1.81 & 5.36 \\ \hline
    \end{tabular}
    \label{tab:xsum-asym-inference-small}
\end{table*}
\begin{table*}[!ht]
    \centering
    \caption{Role of model symmetry in inference efficiency on FLAN-T5 base model on the XSUM dataset}
    \small
    \begin{tabular}{|l|l|l|l|l|l|l|l|l|l|l|l|l|}
    \hline
         $l_{enc}$ & $l_{dec}$ & R-2 & Impact & BS 1 & STD & Speedup & BS 8 & STD  & Speedup & BS 16 & STD & Speedup\\ \hline
         12 & 12 & 15.69 & 0.00\% & 205 & 3.81 & 1.00 & 546 & 8.7 & 1.00 & 917 & 4.72 & 1.00 \\ \hline
        12 & 10 & 15.27 & -2.65\% & 171 & 2.79 & 1.20 & 508 & 6.39 & 1.07 & 876 & 3.02 & 1.05 \\ \hline
        12 & 8 & 14.91 & -4.97\% & 150 & 1.32 & 1.37 & 476 & 2.82 & 1.15 & 830 & 1.08 & 1.10 \\ \hline
        12 & 6 & 15.40 & -1.83\% & 129 & 4.33 & 1.59 & 450 & 9.33 & 1.21 & 789 & 3.73 & 1.16 \\ \hline
        12 & 4 & 15.19 & -3.18\% & 101 & 2.16 & 2.03 & 411 & 5.27 & 1.33 & 744 & 1.71 & 1.23 \\ \hline
        12 & 2 & 13.73 & -12.47\% & 76 & 1.76 & 2.70 & 380 & 3.43 & 1.44 & 706 & 8.13 & 1.30 \\ \hline
        10 & 12 & 15.92 & 1.47\% & 200 & 6.37 & 1.03 & 494 & 2.45 & 1.11 & 818 & 1.72 & 1.12 \\ \hline
        8 & 12 & 14.10 & -10.09\% & 195 & 5.47 & 1.05 & 445 & 20.8 & 1.23 & 713 & 1.71 & 1.29 \\ \hline
        6 & 12 & 13.84 & -11.79\% & 190 & 3.89 & 1.08 & 396 & 9.79 & 1.38 & 612 & 4.64 & 1.50 \\ \hline
        4 & 12 & 12.10 & -22.88\% & 185 & 2.24 & 1.11 & 337 & 3.09 & 1.62 & 505 & 1.96 & 1.82 \\ \hline
        2 & 12 & 10.27 & -34.53\% & 180 & 2.08 & 1.14 & 282 & 4.03 & 1.94 & 399 & 2.85 & 2.30 \\ \hline
        10 & 10 & 15.72 & 0.22\% & 174 & 4.09 & 1.18 & 475 & 18.5 & 1.15 & 772 & 1.79 & 1.19 \\ \hline
        8 & 8 & 14.30 & -8.85\% & 140 & 1.95 & 1.46 & 373 & 2.21 & 1.46 & 625 & 1.51 & 1.47 \\ \hline
        6 & 6 & 12.44 & -20.68\% & 112 & 1.71 & 1.83 & 290 & 6.77 & 1.88 & 480 & 3.5 & 1.91 \\ \hline
        4 & 4 & 10.67 & -31.95\% & 84.2 & 3.75 & 2.43 & 201 & 1.58 & 2.72 & 330 & 4.43 & 2.78 \\ \hline
        2 & 2 & 7.74 & -50.65\% & 51.5 & 3.01 & 3.98 & 112 & 1.02 & 4.88 & 179 & 0.894 & 5.12 \\ \hline
    \end{tabular}
    \label{tab:xswum-asym-inference-base}
\end{table*}

\begin{table*}[!ht]
    \centering
    \caption{Role of model symmetry in inference efficiency on FLAN-T5 large model on the XSUM dataset}
    \small
    \begin{tabular}{|l|l|l|l|l|l|l|l|l|l|l|l|l|}
    \hline
         $l_{enc}$ & $l_{dec}$ & R-2 & Impact & BS 1 & STD & Speedup & BS 8 & STD  & Speedup & BS 16 & STD & Speedup\\ \hline
         24 & 24 & 16.34 & 0.00\% & 447 & 19.4 & 1.00 & 1480 & 23 & 1.00 & 2700 & 16.1 & 1.00 \\ \hline
        24 & 20 & 19.80 & 21.16\% & 374 & 4.84 & 1.20 & 1410 & 17.5 & 1.05 & 2580 & 7.52 & 1.05 \\ \hline
        24 & 16 & 19.30 & 18.09\% & 327 & 19.4 & 1.37 & 1320 & 8.18 & 1.12 & 2460 & 7.19 & 1.10 \\ \hline
        24 & 12 & 18.92 & 15.77\% & 272 & 7.91 & 1.64 & 1240 & 7.06 & 1.19 & 2340 & 7.5 & 1.15 \\ \hline
        24 & 8 & 17.96 & 9.93\% & 216 & 7.81 & 2.07 & 1170 & 11.4 & 1.26 & 2210 & 6.49 & 1.22 \\ \hline
        24 & 4 & 16.47 & 0.76\% & 165 & 3.11 & 2.71 & 1090 & 3.66 & 1.36 & 2080 & 7.17 & 1.30 \\ \hline
        20 & 24 & 19.43 & 18.88\% & 406 & 21.5 & 1.10 & 1310 & 11.5 & 1.13 & 2390 & 7.76 & 1.13 \\ \hline
        16 & 24 & 18.33 & 12.16\% & 412 & 20.3 & 1.08 & 1140 & 6.88 & 1.30 & 2080 & 7.01 & 1.30 \\ \hline
        12 & 24 & 16.90 & 3.39\% & 384 & 18.8 & 1.16 & 986 & 11 & 1.50 & 1750 & 686 & 1.54 \\ \hline
        8 & 24 & 14.97 & -8.39\% & 369 & 8.87 & 1.21 & 822 & 15.5 & 1.80 & 1420 & 15.5 & 1.90 \\ \hline
        4 & 24 & 12.52 & -23.37\% & 345 & 4.41 & 1.30 & 649 & 3.26 & 2.28 & 110 & 5.96 & 24.55 \\ \hline
        20 & 20 & 19.18 & 17.38\% & 357 & 11.8 & 1.25 & 1230 & 13.2 & 1.20 & 2260 & 2.16 & 1.19 \\ \hline
        16 & 16 & 17.56 & 7.43\% & 288 & 5.91 & 1.55 & 995 & 9.41 & 1.49 & 1820 & 5.33 & 1.48 \\ \hline
        12 & 12 & 15.89 & -2.79\% & 217 & 3.09 & 2.06 & 748 & 3.25 & 1.98 & 1370 & 6.59 & 1.97 \\ \hline
        8 & 8 & 12.15 & -25.66\% & 158 & 6.04 & 2.83 & 511 & 9.62 & 2.90 & 920 & 2.06 & 2.93 \\ \hline
        4 & 4 & 8.96 & -45.14\% & 92.3 & 2.88 & 4.84 & 267 & 1.51 & 5.54 & 481 & 1.69 & 5.61 \\ \hline
    \end{tabular}
    \label{tab:xsum-asym-inference-large}
\end{table*}
\subsection{Responsible NLP Research - Reproducibility Checklist}
\subsubsection{Scientific Artifacts}
\noindent\textbf{Datasets.} We perform our experimentation on well-established benchmarks using many broad domains and a proprietary web summarization dataset. We do not perform any modification or augmentation on public benchmarks in any dataset.  \\
\noindent\textbf{Models.} The model used as a starting point for all of our experiments is the family of flan-t5 models, publicly available via HuggingFace Hub~\footnote{https://huggingface.co/bert-base-uncased}. All other models presented in this paper are openly-available in the hugging face hub. 
\subsubsection{Computational Experiments}
Our experimentation on finetuning our compressed models uses a single 40GB A100. Finetuning time varies across datasets ranging from 1 hour for T5-small to 24 hours for T5-Large.  
\subsubsection{Computational Packages}
All of our experimentation is done using public libraries and datasets to ensure extensibility and reproducibility. Our investigation is done using HuggingFace's Transformers \footnote{https://github.com/huggingface/transformers} and Datasets \footnote{https://github.com/huggingface/datasets}.
\end{document}